\documentclass[lettersize,journal]{IEEEtran}
\usepackage{amsmath,amsfonts}
\usepackage{algorithmic}
\usepackage{algorithm}
\usepackage{eqparbox}

\usepackage{array}
\usepackage[caption=false,font=normalsize,labelfont=sf,textfont=sf]{subfig}
\usepackage{textcomp}
\usepackage{stfloats}
\usepackage{url}
\usepackage{verbatim}
\usepackage{graphicx}
\usepackage{cite}
\usepackage[utf8]{inputenc}
\usepackage{amssymb}
\usepackage{amsmath}
\usepackage{graphicx}  
\usepackage{xcolor}
\DeclareMathOperator*{\argmax}{arg\,max}

\usepackage{tabularx}
\usepackage{color}
\usepackage{wrapfig}    
\usepackage{booktabs}
\usepackage{dsfont}
\usepackage{bbm}
\usepackage{caption}
\usepackage{multirow}
\usepackage[font=footnotesize]{caption} 
\usepackage{amsmath,amssymb,amsbsy}
\usepackage[english]{babel}
\usepackage{amsthm}
\newtheorem{theorem}{Theorem}
\newtheorem{lemma}{Lemma}
\usepackage[hidelinks]{hyperref}       
\usepackage{indentfirst}
\usepackage{orcidlink}

\usepackage{color}

\begin{document}

\title{Value Approximation for Two-Player General-Sum Differential Games with State Constraints}

\author{Lei Zhang$^{\orcidlink{0009-0004-3023-0306}}$,~\IEEEmembership{Student Member,~IEEE,} Mukesh Ghimire$^{\orcidlink{0009-0005-5660-5054}}$,~\IEEEmembership{Student Member,~IEEE,} Wenlong Zhang$^{\orcidlink{0000-0002-4046-2213}}$,~\IEEEmembership{Member,~IEEE,} Zhe Xu$^{\orcidlink{0000-0002-0440-0912}}$,~\IEEEmembership{Member,~IEEE,} Yi Ren$^{\orcidlink{0000-0003-2306-5993}}$,~\IEEEmembership{Member,~IEEE} 
\thanks{Lei Zhang, Mukesh Ghimire, Zhe Xu, and Yi Ren are with Department of Mechanical and Aerospace Engineering, Arizona State University, Tempe, AZ, 85287, USA. Email:{\tt\small \{lzhan300, mghimire, xzhe1, yiren\}@asu.edu}.}
\thanks{Wenlong Zhang is with School of Manufacturing Systems and Networks, Ira A. Fulton Schools of Engineering, Arizona State University, Mesa, AZ, 85212, USA. Email:{\tt\small wenlong.zhang@asu.edu}.}%
\thanks{Yi Ren is the corresponding author.}%
}

\markboth{IEEE TRANSACTIONS ON ROBOTICS}%
{Shell \MakeLowercase{\textit{et al.}}: A Sample Article Using IEEEtran.cls for IEEE Journals}


\maketitle

\begin{abstract}
Solving Hamilton-Jacobi-Isaacs (HJI) PDEs numerically enables equilibrial feedback control in two-player differential games, yet faces the curse of dimensionality (CoD). While physics-informed neural networks (PINNs) have shown promise in alleviating CoD in solving PDEs, vanilla PINNs fall short in learning discontinuous solutions due to their sampling nature, leading to poor safety performance of the resulting policies when values are discontinuous due to state or temporal logic constraints. In this study, we explore three potential solutions to this challenge: (1) a hybrid learning method that is guided by both supervisory equilibria and the HJI PDE, (2) a value-hardening method where a sequence of HJIs are solved with increasing Lipschitz constant on the constraint violation penalty, and (3) the epigraphical technique that lifts the value to a higher dimensional state space where it becomes continuous. Evaluations through 5D and 9D vehicle and 13D drone simulations reveal that the hybrid method outperforms others in terms of generalization and safety performance by taking advantage of both the supervisory equilibrium values and costates, and the low cost of PINN loss gradients.


\end{abstract}

\begin{IEEEkeywords}
general-sum differential game, physics-informed neural network, safe human-robot interactions 
\end{IEEEkeywords}


\section{Introduction}
\label{sec:intro}
\IEEEPARstart{H}{uman-robot} interactions (HRI) become prevalent in safety-critical applications such as transportation~\cite{duarte2018impact}, healthcare~\cite{peters2018review}, and rescue~\cite{murphy2004human}. 
Conventionally, safety is achieved by incorporating state constraints in a model predictive control (MPC) framework. The constraints are usually derived from a two-player zero-sum game formulation so that the ego player avoids all system states from which the fellow player can successfully launch attacks should it be adversarial~\cite{leung2020infusing}. 
There are two limitations to this approach: First, the zero-sum setting can often be overly conservative since fellow players in civil applications are not always adversarial; Second, real-time MPC is required on top of value approximation of the zero-sum games, limiting the speed and quality of the player's decision making.

To address the first limitation, it is tempting to consider HRI as general-sum differential games with state constraints and incomplete information, where players have private types (e.g., reward parameters). In this setting, players can overcome unnecessary conservatism by updating their beliefs about each other's type based on observations of their previous actions.
To address the second limitation, one would ideally need to obtain the value of the game, which then enables feedback control that intrinsically satisfies the state constraints while optimizing the expected payoff, either obsoleting or at least accelerating MPC. 

A theoretical challenge towards these idealistic goals, however, is that we do not have the existence proof or the characterization of values for general-sum differential games with incomplete information and state constraints~\cite{gammoudi2023differential}.
Hence we take a step back and consider games with complete information, for which Nash equilibrium exists~\cite{lewis2012optimal} and therefore values are governed by the Hamilton-Jacobi-Isaacs (HJI) equations. Computing values, however, is known to encounter the curse of dimensionality (CoD) using mesh-based dynamic programming (DP) solvers~\cite{bellman1965dynamic}.
Physics-informed neural network (PINN) has thus been introduced to approximate values while circumventing CoD~\cite{weinan2021algorithms}. Nonetheless, recent studies showed that while PINN is successful at approximating Lipschitz continuous PDE solutions~\cite{weinan2021algorithms,shin2020convergence,ito2021neural}, they encounter convergence issues when applied to discontinuous ones~\cite{fuks2020limitations}. In the context of HJI, such value discontinuity arises when state constraints and temporal logic specifications are imposed.

Within this context, our paper investigates three PINN-based solutions for approximating values of state-constrained differential games: 

The first solution, called \textit{hybrid learning}, is developed based on the insight that discontinuity in value causes sampling-based methods such as PINN to deviate from the true solutions almost surely, since the measure of the discontinuous boundaries is zero (or close to zero when we approximate discontinuities with large-Lipschitz functions in practice). The solution is thus to augment PINN with supervised equilibrium data that cover discontinuous regions of the value landscape in space and time. These equilibria are generated by solving boundary value problems (BVPs) following Pontryagin's Maximum Principle (PMP)~\cite{mangasarian1966sufficient}. This solution requires human insights on where the informative equilibrium trajectories with discontinuous values (e.g., collisions) lie and the global optimality of the BVP solutions. 
The challenge with sampling discontinuous boundaries leads to the loss of spatiotemporal causality during value approximation. Hence the second solution, called \textit{value hardening}, following curriculum learning~\cite{bengio2009curriculum}, aims to improve the chance of learning the discontinuous boundaries by gradually increasing the Lipschitz constant of a constraint violation penalty.
The third solution, called \textit{epigraphical learning}, is based on the epigraphical technique that transforms discontinuous values of state-constrained games into Lipschitz continuous ones defined in an augmented state space~\cite{altarovici2013general}. We extend the existing technique from zero-sum games~\cite{lee2022safety} to the general-sum setting and apply PINN to approximate the smooth augmented values.



We summarize the systemic design of experiments to be used to evaluate and compare these solutions:
\textbf{Methods:} vanilla PINN, supervised learning (SL), hybrid (HL), value-hardening (VH), epigraphical (EL) methods and their variants. \textbf{PDE dimensionalities:} 5D, 9D, and 13D. As of writing, 13D is largest dimensionality among existing test cases of HJ equations in the differential game context. \textbf{Dynamics:} linear and nonlinear vehicle and drone dynamics. \textbf{Information settings:} complete- and incomplete-information two-player general-sum games. \textbf{Performance metrics:} both in- and out-of-distribution generalization and safety performance.

We claim the following contributions: 

\begin{itemize}
    \item We show that HL scales better than SL, VH and EL to high-dimensional cases in terms of both generalization (value and action prediction) and safety (when values are used for feedback control). The key factors for its success are (1) the supervision on the costate landscape, which is directly related to the control policy, and (2) the low cost of PINN training in comparison to supervised learning via solving BVPs.
    \item Consistent with \cite{raissi2019physics,raissi2019deep,jagtap2020adaptive}, our ablation studies highlight the sensitivity of generalization and safety performance to the choice of neural activation functions, and the need for adaptive activations. 
    In particular, \texttt{tanh} and continuously differentiable variants of \texttt{relu}, such as \texttt{gelu}~\cite{hendrycks2016gaussian}, achieve the best empirical performance when combined with HL and adaptive activation.
    \item While existing studies on solving HJ equations using machine learning have shown promising results for reachability analysis (e.g., \cite{deepreach}), the safety performance of the resultant value networks when used as closed-loop controllers is rarely investigated. We show in this paper that low approximation errors in value does not necessarily indicate high safety performance when the approximated value is used for closed-loop control. 
\end{itemize}

This work is extended from its conference version~\cite{zhang2023approximating} in the following significant ways: (1) a thorough investigation of the efficacy of the epigraphical technique when applied to solving differential games; (2) new studies to demonstrate and explain the convergence challenge encountered when applying value hardening to 9D and 13D problems; (3) 
new studies that demonstrate the importance of costate loss for high safety performance;
(4) extension of the existing DP solver~\cite{bui2022optimizeddp} from zero-sum to general-sum setting, which enables comparisons between values approximated by DP, BVP, and PINN variants. These comparisons allow us to show that values obtained from all three are similar in the test cases and therefore guiding PINN by open-loop BVP solutions through hybrid learning is reasonable.


The rest of this paper is organized as follows. Sec.~\ref{sec:relate work} provides an overview of the relevant literature on value approximation, physics-informed neural network, and complete- and incomplete-information differential games. In Sec.~\ref{sec:method}, we present the formulation of two-player general-sum differential games with state constraints, its HJ PDEs, and explain the challenge in approximating its discontinuous values through a toy case. We then discuss the three potential solutions. The experimental results are presented and analyzed in Sec.~\ref{sec:case}. We give discussion including safety guarantee, consistency between BVP and HJI values, and efficacy of costate loss for safety performance in Sec.~\ref{sec:discussion}, and conclude in Sec.~\ref{sec:conclusion}.

\vspace{-0.1in}
\section{Related Work}
\label{sec:relate work}
\subsection{Value approximation and physics-informed neural network}
The values of a general-sum differential game with two-player and complete-information are viscosity solutions to HJI equations~\cite{crandall1983viscosity}, which are a set of first-order nonlinear PDEs. 
The conventional approach to solving such equations involves essentially non-oscillatory (ENO) schemes~\cite{osher1991high} and level set methods~\cite{osher2004level, mitchell2005toolbox}, which are known to provide accurate approximations of both temporal and spatial derivatives.   
However, these approaches suffer from CoD~\cite{mitchell2003}. 
Recent studies have shown that using physics-informed neural network (PINN) to approximate PDE solutions can effectively circumvent the CoD due to its Monte Carlo nature, provided that the solution is smooth~\cite{weinan2021algorithms}. 
PINN trains neural nets as PDE-governed fields, where the training loss is defined by network-induced residuals with respect to (a) the boundary conditions~\cite{han2020convergence, han2018solving}, (b) the governing equations~\cite{jagtap2020adaptive,deepreach}, and/or (c) supervisory data drawn from the ground-truth solutions~\cite{nakamura2021adaptive}.
Initial studies on convergence and generalization performance have emerged for (a) and (b), under the assumption that both the solution and the network are Lipschitz continuous~\cite{han2020convergence,shin2020convergence,ito2021neural}. Recent studies have explored the effectiveness of PINN for solving PDEs with discontinuous solutions, such as Burgers' equation, where both initial and terminal boundaries are specified~\cite{jagtap2020adaptive}. However, we demonstrate in Sec.~\ref{sec:challenge_piml} that PDEs with only terminal or initial boundary conditions, such as HJIs, present an unidentifiability challenge.

\vspace{-0.1in}    
\subsection{Differential games with incomplete information} 
One driving motivation for approximating values of differential games is to use the values for fast belief updates on unknown player types in incomplete-information settings. The update follows Bayes inference and relies on modeling player control policies as a type-conditioned distribution shaped by their values (see Sec.~\ref{sec:uncontrolled intersection} for details). 
In the case study on uncontrolled intersection (Sec.~\ref{sec:uncontrolled intersection}), we evaluate the safety performance induced by the value networks, which influence both players' control policies and their belief updates about the types of their fellow players. In addition, we examine safety performance when players are ``empathetic'', i.e., when they share common beliefs about each other, and when they are ``non-empathetic'', i.e., when they falsely assume that their true types are known by their fellow players. 
Our study shares the same motivation as \cite{fridovich2020efficient} in that both seek fast computation of equilibrium during interactions. We take the approach of pre-computing values offline (which then enables 500Hz policy generation frequency during inference time), while \cite{fridovich2020efficient} proposes to simplify games as linear-quadratic which then facilitates fast (20Hz) equilibrium approximation online.
Our investigation into differential games with incomplete information sets us apart from previous HRI studies that resort to various simplifications of the games in order to balance theoretical soundness and practicality. These simplifications involve modeling the games as optimal control problems or complete-information ones~\cite{foerster_learning_2017,sadigh_planning_2018,kwon2020humans,schwarting2019social,zahedi2022seeking,li2022off}. While some also use belief updates to adapt motion planning, they are limited to empirical best responses of the uninformed player in one-sided information settings~\cite{nikolaidis2017human,sun2018probabilistic,peng2019bayesian,wang2019enabling,fridovich2020confidence}. A recent study proposes to synthesize safety control policies that account for evolving uncertainty by considering both physical and belief dynamics~\cite{hu2023learning}. This framework is currently constrained to one-sided information settings, while this paper studies cases where both players lack information. 
It is necessary to point out, however, that we will only investigate best-response policies of players, i.e., the players choose the best responses based on their \textit{current} belief about their fellows (via their common knowledge about the values of the games) without considering the \textit{future} dynamics of beliefs. This is because the existence of value and player policies for general-sum differential games with incomplete information is still an open question, unlike their zero-sum or discrete-time counterparts~\cite{aumann,cardaliaguet2007differential, cardaliaguet2009numerical, cardaliaguet2012games}.

\vspace{-0.1in}
\section{Discontinuous Value Approximation}
\label{sec:method}
\subsection{Notations}
In a two-player differential game with {complete-information}, Player $i$ has a state space $\mathcal{X}_i \subset \mathbb{R}^n$ and an action space $\mathcal{U}_i \subset \mathbb{R}^m$. The time-invariant state dynamics of Player $i$ is denoted by 
\begin{equation}
    \dot{x}_i = f_i(x_i, u_i),
    \label{eq:dynamics}
\end{equation}
where $x_i\in\mathcal{X}_i$ and $u_i\in\mathcal{U}_i$. We omit dependence on time whenever possible and use $\textbf{a}_i=(a_i, a_{-i})$ to concatenate variables $a_i$ from Player $i$ and $a_{-i}$ from the fellow player. We denote the partial derivative with respect to $x$ by $\nabla_x\cdot$ and the joint state space by $\mathcal{X} := \bigcup_{i=1,2}\mathcal{X}_i$. The fixed time horizon of the game is $[0, T]$.
The instantaneous loss of Player $i$ is denoted by $l_i(\textbf{x}_i, u_i)$ and the terminal loss $g_i(x_i)$. Feasible states from Player $i$'s perspective are defined by the sub-zero level set $\{\textbf{x}_i \in \mathcal{X} ~|~ c_i(\textbf{x}_i) \leq 0\}$. We will consider $c_i(\cdot)$ a scalar function that measures the worse state constraint violation in case multiple constraints are present, i.e., if $c_i(\textbf{x}_i) > 0$, $\textbf{x}_i$ violates at least one of the constraints.
The value function of Player $i$ is denoted by $\vartheta_i(\textbf{x}_i, t): \mathcal{X} \times [0,T] \rightarrow \mathbb{R}$. To simplify notation, we will use $f_i$, $l_i$, $g_i$, $c_i$, and $\vartheta_i$ to refer to the dynamics, losses, state constraint, and the value function of Player $i$.
Denote by $\alpha_i \in \mathcal{A}: \mathcal{X} \times [0,T] \rightarrow \mathcal{U}_i$ Player $i$'s control policy, where the policy space $\mathcal{A}$ is assumed to be common. We use $x_s^{x_i,t,\alpha_i}$ as the state of Player $i$ at time $s$ if it follows policy $\alpha_i$ and dynamics $f_i$ starting from $(x_i,t)$. We denote states for two players at time $s$ as $\textbf{x}_s^{\textbf{x}_i,t,\alpha_i,\alpha_{-i}}:=\left(x_s^{x_i,t,\alpha_i}, x_s^{x_{-i},t,\alpha_{-i}}\right)$. All acronyms are summarized in Sec.~\ref{sec:acronym}.

\vspace{-0.1in} 
\subsection{Assumptions}
\label{sec:assumptions}
Throughout the paper, we assume that $\mathcal{U}_i$ is compact and convex; $f_i: \mathcal{X}_i \times \mathcal{U}_i \rightarrow \mathbb{R}^n$ and $c_i: \mathcal{X} \rightarrow \mathbb{R}$ are Lipschitz continuous; 
$l_i: \mathcal{X}_i  \times \mathcal{U}_i \rightarrow \mathbb{R}$ and $g_i: \mathcal{X}_i \rightarrow \mathbb{R}$ are Lipschitz continuous and bounded.


\vspace{-0.1in} 
\subsection{Preliminary}
\label{sec:preliminary}
\textit{Hamilton-Jacobi-Isaacs equations}: Let $(\alpha_i, \alpha_{-i})$ be a pair of equilibrium policies. The values for a two-player general-sum differential game are viscosity solutions to the HJI equations denoted by ($L$) in Eq.~\eqref{eq:hji}, and satisfy the boundary conditions denoted by ($D$)~\cite{starr1969nonzero}.
\begin{equation}
\begin{aligned}
& L(\vartheta_i, \nabla_{\textbf{x}_i} \vartheta_i, \textbf{x}_i, t, \xi_{-i}) := \nabla_t \vartheta_i + \max_{u_i \in \mathcal{U}_i} \left\{\nabla_{\textbf{x}_i} \vartheta_i ^T \textbf{f}_i - l_i\right\} = 0 \\
& D(\vartheta_i, \textbf{x}_i) := \vartheta_i(\textbf{x}_i, T) - g_i = 0, \quad  \text{for} \quad i = 1, 2. \\
\end{aligned}
\label{eq:hji}
\end{equation}
With the values, the players' equilibrium policies can then be derived by $\alpha_i(\textbf{x}_i, t) = \argmax_{u_i \in \mathcal{U}_i} \left\{\nabla_{\textbf{x}_i} \vartheta_i^T \textbf{f}_i - l_i \right\}$. Notice that $L$ for Player $i$ depends on the equilibrium policy $\alpha_{-i}$ of its fellow. 

\textit{Pontryagin's Maximum Principle}: 
Although solving the HJI equations would give a feedback control policy, it is often more practical to compute open-loop policies for a specific initial state $(\bar{x}_1, \bar{x}_2) \in \mathcal{X}$ by solving a boundary value problem (BVP) following Pontryagin's Minimum Principle (PMP)\footnote{It should be noted that solving the BVP has its own numerical challenges, particularly when the equilibrium involves singular arcs~\cite{singarcs}. However, these challenges are beyond the scope of this paper.}:
\begin{equation}
\begin{aligned}
& \dot{x}_i = f_i, ~\quad x_i(0) = \bar{x}_i, \\
& \dot{\lambda}_i = - \nabla_{x_i} h_i, ~\quad \lambda_i(T) = - \nabla_{x_i} g_i, \\
& u_i = \argmax_{u_i \in \mathcal{U}_i} ~\{h_i\}  \quad \text{for} \quad i = 1, 2.
\end{aligned}
\label{eq:pmp}
\end{equation}
Here $\lambda_i$ is the time-dependent co-state for Player $i$. The co-state connects PMP and HJI through $\lambda_i = \nabla_{x_i} \vartheta_i$. Solutions to Eq.~\eqref{eq:pmp} are specific to the given initial states. Although PMP characterizes local open-loop solutions, empirical studies (see Sec.~\ref{sec:numerical value comparison}) show that with an effort to search for global solutions, BVP values are consistent with those governed by the HJI equations.

\textit{State-constrained value function}:
With state constraints, the value function for Player $i$ with some equilibrium policy pair $(\alpha_i, \alpha_{-i})$ is 
\begin{equation}
    \vartheta_i(\textbf{x}_i, t) = \int_t^T l_i\left(x_s^{\textbf{x}_i,t,\alpha_i}, \alpha_i\left(\textbf{x}_s^{\textbf{x}_i,t,\alpha_i,\alpha_{-i}}, s\right)\right)ds + g_i\left(x_T^{\textbf{x}_i,t,\alpha_i}\right),
\label{eq:value_fn}    
\end{equation}
if $c_i\left(\textbf{x}_s^{\textbf{x}_i,t,\alpha_i,\alpha_{-i}}\right) \leq 0, \forall s \in [t, T]$, or $+\infty$ otherwise. Thus state constraints introduce discontinuity in the value landscape.

\vspace{-0.1in}
\subsection{PINN for solving HJ equation}
PINN trains neural networks $\hat{\vartheta}_i(\cdot,\cdot): \mathcal{X} \times [0,T] \rightarrow \mathbb{R}$ to approximate $\vartheta_i$. We denote by $\mathcal{D} = \left \{\left(x_1^{(k)}, x_2^{(k)}, t^{(k)} \right) \right\}_{k=1}^K$ a dataset consisting of uniform samples in $\mathcal{X}_1 \times \mathcal{X}_2 \times [0, T]$. The formulation of the training problem in Eq.~\eqref{eq:deepreach} extends from PINN for solving zero-sum games~\cite{deepreach}:
\begin{equation}
\begin{aligned}
    \min_{\hat{\vartheta}_1, \hat{\vartheta}_2} \quad L_1\left(\hat{\vartheta}_1, \hat{\vartheta}_2; \theta \right) := & \sum_{k=1}^K \sum_{i=1}^2 \left\|L(\hat{\vartheta}_i^{(k)}, \nabla_{\textbf{x}_i} \hat{\vartheta}_i^{(k)}, \textbf{x}_i^{(k)}, t^{(k)})\right\|\\
    & + C_1 \phi\left(D(\hat{\vartheta}_i, \textbf{x}_i^{(k)})\right),
    \label{eq:deepreach}
\end{aligned}
\end{equation}
where $\hat{\vartheta}_i^{(k)}$ is an abbreviation for $\hat{\vartheta}_i \left(\textbf{x}_i^{(k)}, t^{(k)} \right)$ and $C_1$ balances the L1 PDE residual loss ($\|L\|$) and the boundary loss ($\phi(D)$). It is worth noting that in each iteration of solving Eq.~\eqref{eq:deepreach}, a sub-routine is needed to find the control policies by maximizing the Hamiltonian. 

\begin{figure}[!ht]
\centering
\vspace{-10pt}
\includegraphics[width = 0.96\linewidth]{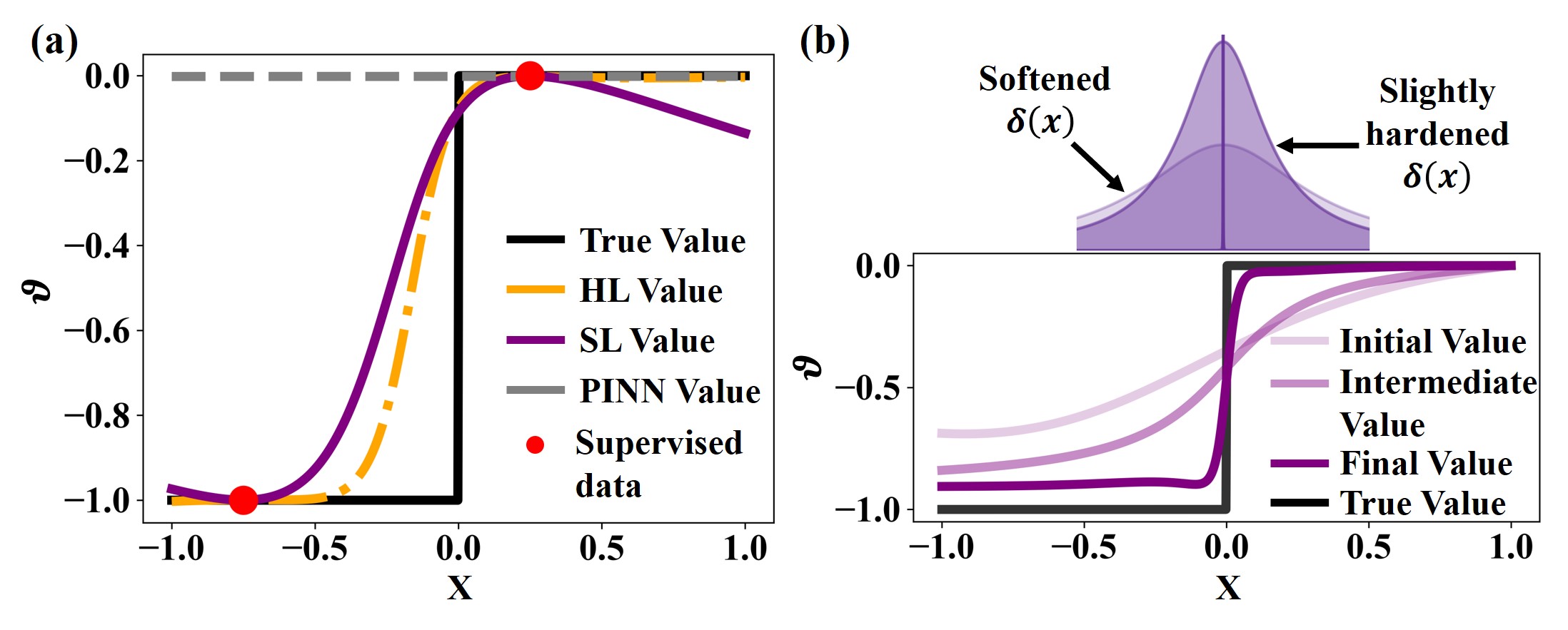}
\caption{(a) Value comparison among the learning methods for a simple 1D case. Red dots are the supervised data. (b) Evolution of the value function due to gradually hardening delta function. Delta functions are shown on top. Transparency reduces with hardening.}
\label{fig:toy_case}
\vspace{-15pt}
\end{figure}

\vspace{-0.1in} 
\subsection{Challenge in approximating discontinuous HJI values}
\label{sec:challenge_piml}
We use the following toy case to explain the challenge in approximating discontinuous values using PINN. Consider a one-dimensional function $\vartheta(x)$, which is the solution to a differential equation $\nabla_x \vartheta - \delta(x) = 0$ with the boundary condition $\vartheta(1) = 0$ in the interval $x \in [-1, 1]$. $\delta(x)$ is a delta function that peaks at $x = 0$. Notice that with uniform samples for $\mathcal{D}$, the PINN loss ($L_1$) can be minimized almost surely by incorrect solutions, e.g., $\hat{\vartheta}(x) = 0$. This unidentifiability issue is due to the differential nature of the governing equation: the accuracy of $\hat{\vartheta}$ at one point in space and time depends solely on that of its neighbors. However, informative neighbors, i.e., those at $x = 0$ in this toy case, have zero probability to be sampled. 

\vspace{-0.1in} 
\subsection{Solutions}
(1) \textit{Hybrid learning}: 
In the above toy case, we can learn a much improved approximation to the solution using only two informative data points sampled from each side of $0$ (as shown by the SL curve in Fig.~\ref{fig:toy_case}). Indeed, \cite{nakamura2021adaptive} showed that supervised learning can be used for value approximation. A drawback of this approach, when applied to solving HJIs, is its high data acquisition costs due to the need for repeatedly solving BVPs to acquire state-value pairs. We hypothesize that this drawback can be reduced by combining supervised learning and PINN, since evaluating and differentiating the latter only require one forward pass of $\hat{\vartheta}$, which is usually much cheaper than calling the Newton-type iterative algorithms involved in solving BVPs.

To implement this hybrid method, we define a dataset $\mathcal{D}_s = \left \{\left(\textbf{x}_i^{(k)}, t^{(k)}, \vartheta_i^{(k)}, \nabla_{\textbf{x}_i}\vartheta_i^{(k)} \right) \text{ for } i=1,2 \right\}_{k=1}^K$ derived from solving Eq.~\eqref{eq:pmp} with initial states uniformly sampled in $\mathcal{X}$. We define the supervised loss as follows:
\begin{equation}
\begin{aligned}
    \min_{\hat{\vartheta}_1, \hat{\vartheta}_2} \quad L_2\left(\hat{\vartheta}_1, \hat{\vartheta}_2; \mathcal{D}_s\right) := & \sum_{k=1}^K \sum_{i=1}^2 \left| \hat{\vartheta}_i^{(k)} - \vartheta_i^{(k)} \right|\\
    & + C_2 \left \|\nabla_{\textbf{x}_i}\hat{\vartheta}_i^{(k)} - \nabla_{\textbf{x}_i} \vartheta_i^{(k)}\right \|,
    \label{eq:supervised}
\end{aligned}
\end{equation}
where $C_2$ is a hyperparameter that balances the losses on value and its gradient. The hybrid method minimizes $L_1 + L_2$.

(2) \textit{Value hardening}: The second solution is to introduce a surrogate differential equation, which has a continuous solution that approximates the ground truth. We can then approximate the true solution by gradually ``hardening'' this surrogate. For the toy case, we can improve the solution by gradually hardening a softened delta function, as shown in Fig.~\ref{fig:toy_case}b. Just like hybrid learning, this method also introduces additional computation, as we turn one learning problem into a sequence of easier learning problems.  
In Sec.~\ref{sec:case}, we show that with a limited budget, value hardening fails to converge for high-dimensional value approximation tasks where hybrid learning succeeds. Lastly, we note that value hardening is similar to~\cite{fuks2020limitations}, where the authors introduce a gradually hardening diffusion term to address the same discontinuity issue when solving nonlinear two-phase hyperbolic transport equations using PINN.  

(3) \textit{Epigraphical learning}: Recall that the discontinuity of value in our context is caused by state constraints in differential games. It is shown that a smooth augmented value can be derived through the epigraphical technique for state-constrained differential games~\cite{altarovici2013general,lee2022safety}. Our last approach utilizes this technique to facilitate continuous value approximation in an augmented state space and compute the value for the original game based on the approximation. While HJ PDEs with state constraint have been investigated in zero-sum settings and numerical approximation of their values have been attempted via dynamic programming and conservative Q-learning~\cite{lee2022safety,kumar2020conservative}, this paper is among the first to solve general-sum differential games with state constraints using a combination of PINN and the epigraphical technique. For completeness, we briefly introduce the epigraphical technique in the following subsection.

\vspace{-0.2in}
\subsection{The epigraphical technique for general-sum differential games with state constraints}

Let $(\alpha_1,\alpha_2)$ be a pair of equilibrium policies. The epigraphical technique introduces an augmented value $V_i: \mathcal{X} \times \mathbb{R} \times [0,T]$:
\begin{equation}
\begin{aligned}
V_i(\textbf{x}_i,z_i,t): = \max\bigg\{\max_{s\in [t,T]}c_i\left(\textbf{x}_s^{\textbf{x}_i,t,\alpha_i,\alpha_{-i}}\right), \\
g_i\left(x_T^{\textbf{x}_i,t,\alpha_i}\right) - z_i(T)\bigg\}
\label{eq:big v_fn_g}
\end{aligned}
\end{equation}
The auxiliary state $z_i$ follows 
\begin{equation}
\dot z_i = -l_i({x}_i, u_i) \ \text{and} \ z_i(0) = \bar z_i,
\label{eq:z_fn_g}
\end{equation}
where $\bar{z}_i$ represents the true value of Player $i$ at $(\bar{\textbf{x}}_i, t_0) \in \mathcal{X} \times [0,T]$ and is computed as follows: Find $\bar z_i \in [z_{min}, z_{max}]$ such that $V_i(\bar{\textbf{x}}_i,\bar z_i,t_0) = 0$. If $V_i(\bar{\textbf{x}}_i,z,t_0) > 0$ for all $z \in [z_{min}, z_{max}]$, then $\bar{z}_i = +\infty$. Lemma~\ref{lemma:lemma1} (Lemma 1 of \cite{lee2022safety}) formally establishes this connection between the augmented value $V_i$ and the true value $\vartheta_i(\textbf{x}_i, t)$:

\begin{lemma}
\label{lemma:lemma1}
Suppose assumptions in Sec.~\ref{sec:assumptions} hold. For all $(\textbf{\emph x}_i,z_i,t) \in \mathcal{X} \times \mathbb{R} \times [0, T]$, $\vartheta_i$ and $V_i$ are related as follows:
\begin{equation}
\begin{aligned}
& \vartheta_i(\textbf{\emph x}_i, t)-z_i \leq 0 \iff V_i(\textbf{\emph x}_i, z_i, t)\leq 0; \\
& \vartheta_i(\textbf{\emph x}_i, t) = \min z_i \ \quad \text{s.t.}\ V_i(\textbf{\emph x}_i,z_i,t)\leq 0.
\end{aligned}
\end{equation}
\end{lemma}

\noindent
\textbf{Proof.} See Sec.~\ref{sec:Lemma 1}.

Lemma~\ref{lemma:lemma2} (Lemma 2 of \cite{lee2022safety}) provides the optimality condition for $V_i(\textbf{x}_i,z_i,t)$, which is the basis for the derivation of HJ equations with state constraints.

\begin{lemma}
\label{lemma:lemma2}
Suppose assumptions in Sec.~\ref{sec:assumptions} hold. For all $(\textbf{\emph x}_i,z_i,t) \in \mathcal{X} \times \mathbb{R} \times [0, T]$, for small enough $h>0$ such that $t+h \leq T$ we have 
\begin{equation}
\begin{aligned}
V_i(\textbf{\emph x}_i,z_i,t) = 
\min_{\alpha_i \in \mathcal{A}}\max\bigg\{\max_{s\in [t,t+h]}c_i\left(\textbf{\emph x}_s^{\textbf{\emph x}_i,t,\alpha_i,\alpha_{-i}}\right), \\
V_i\left(\textbf{\emph x}_i(t+h), z_i(t+h), t+h\right)\bigg\},
\end{aligned}
\end{equation}
where $\textbf{\emph x}_s^{\textbf{\emph x}_i,t,\alpha_i,\alpha_{-i}}$ and $\textbf{\emph x}_i(t+h)$ are solutions to Eq.~\eqref{eq:dynamics} using $(\textbf{\emph x}_i, t, u_i)$ and $z_i(t+h)$ is a solution to Eq.~\eqref{eq:z_fn_g}. $\alpha_{-i}$ is the equilibrium policy of the fellow player.
\end{lemma}
\noindent
\textbf{Proof.} See Sec.~\ref{sec:Lemma 2}.

Theorem~\ref{theorem:theorm1} presents the HJ equations for players in a general-sum differential game with state constraints: 
\begin{theorem}[HJ PDE with state constraints for general-sum differential games]
\label{theorem:theorm1}
For all $(\textbf{\emph x}_i,z_i,t) \in \mathcal{X} \times \mathbb{R} \times [0, T]$, $V_i(\textbf{\emph x}_i,z_i,t)$ in Eq.~\eqref{eq:big v_fn_g} is a viscosity solution to the following HJ PDE and boundary conditions:
\begin{equation}
\begin{aligned}
\max\{&c_i\left(\textbf{\emph x}_i\right)-V_i(\textbf{\emph x}_i,z_i,t), \\
& \nabla_{t} V_i - \mathcal{H}_i(\textbf{\emph x}_i,z_i,\nabla_{\textbf{\emph x}_i}{V}_i,\nabla_{z_i} V_i, t)\} = 0,
\end{aligned}
\end{equation}
where $\mathcal{H}_i$ is the augmented Hamiltonian:
\begin{equation}
\begin{aligned}
\mathcal{H}_i = \max_{u_i \in \mathcal{U}_i}-\nabla_{\textbf{\emph x}_i}{V}_i^T\textbf{\emph f}_i+\nabla_{z_i} V_i^Tl_i,
\label{eq:hjpde_ham}
\end{aligned}
\end{equation}
and $V_i(\textbf{\emph x}_i, z_i, T)=\max\left\{c_i(\textbf{\emph x}_i), g_i(T) - z_i(T)\right\}$.
\end{theorem}

\noindent
\textbf{Proof.} See Sec.~\ref{sec:Theorem 1}.

To solve state-constrained HJ PDEs using PINN, we define residuals similar to Eq.~\eqref{eq:hji}. 
\begin{equation}
\begin{aligned}
& \tilde {L}(V_i, \textbf{x}_i,z_i,t) := \max\big\{c_i\left(\textbf{x}_i\right)-V_i(\textbf{x}_i,z_i,t), \\
& \nabla_{t} V_i - \mathcal{H}_i(\textbf{x}_i,z_i,\nabla_{\textbf{x}_i}{V}_i,\nabla_{z_i} V_i, t)\big\} \\
& \tilde {D}(V_i, \textbf{x}_i, z_i) := V_i(\textbf{x}_i, z_i, T) - \max\big\{c_i(\textbf{x}_i), \\
& g_i(T) - z_i(T)\big\}, \ \text{for} \ i = 1, 2. \\
\end{aligned}
\label{eq:hjpde_constraint}
\end{equation}

Thus, the overall loss can be expressed using the same formulation as in Eq.~\eqref{eq:deepreach}.
\begin{equation}
\begin{aligned}
    \min_{\hat{V}_1, \hat{V}_2} \quad L_3\left(\hat{V}_1, \hat{V}_2; \theta \right) := & \sum_{k=1}^K \sum_{i=1}^2 \left\|\tilde {L}(\hat{V}_i^{(k)}, \textbf{x}_i^{(k)}, z_i^{(k)}, t^{(k)})\right\|\\
    & + C_3 \tilde {\phi}\left(\tilde {D}(\hat{V}_i^{(k)}, \textbf{x}_i^{(k)},z_i^{(k)})\right),
    \label{eq:hjpde_loss}
\end{aligned}
\end{equation}

\noindent
To take advantage of the structure of $V_i$, we introduce two networks $A_i: \mathcal{X} \times [0,T] \rightarrow \mathbb{R}$ and $B_i: \mathcal{X} \times [0,T] \rightarrow \mathbb{R}$:
\begin{equation}
\hat{V}_i(\textbf{x}_i,z_i,t):= \max\left\{A_i(\textbf{x}_i, t), B_i(\textbf{x}_i, t) - z_i\right\}.
\end{equation}
Essentially, $A_i$ predicts the worse-case future constraint violation and $B_i$ predicts the value of the game for Player $i$ without considering the constraint. If $A_i > 0$, then $\hat{V}_i >0$ and $\vartheta$ does not exist, i.e., state constraint cannot be satisfied.

\vspace{-0.1in}
\section{Case Study}
\label{sec:case}
We conduct empirical studies to compare the generalization and safety performance of value approximation models using five different learning methods: vanilla PINN (shortened as PINN), hybrid learning (HL), value hardening (VH), epigraphical learning (EL), and supervised learning (SL). We use both vehicle and drone simulations to formulate the games. The first simulation involves an interaction between two players (i.e., vehicle) at an uncontrolled intersection, which leads to HJIs with coupled value functions defined on a 5D state space. We study both complete- and incomplete-information settings using this simulation. The second and third studies investigate model safety performance on a 9D state space. The former models a collision-avoidance case and the latter a double-lane change case. It should be noted that our settings, in terms of the dynamical models and the state space dimensions, are similar to those of \cite{deepreach} and \cite{leung2020infusing}, yet we extend from their optimal control or zero-sum settings to general-sum differential games. The last case study on drone collision avoidance investigates performance of PINN variants on a higher dimensional state space (13D) and on nonlinear dynamics.

\textbf{Data acquisition.} 
The methods under comparison involve diverse data acquisition algorithms (supervised data via iterative BVP solving and PINN data via random sampling) and learning algorithms (supervised and curriculum learning). Hence, we use the total wall-clock time for data acquisition and learning as a unified measure of the computational cost. To ensure a fair comparison, the data size for each method is chosen to keep their computational costs as close to each other as possible. Computational costs of all the case studies are summarized in Tab.~\ref{table:time cost}. To improve training convergence, we normalize the input data to lie in $[-1, 1]$.

\begin{table}[!ht]
\vspace{-5pt}
\scriptsize
\centering
\caption{Computational costs for all learning methods in all case studies.}
\label{table:time cost}
\begin{tabular}{ccccccc}
\toprule
\multirow{2}{*}{Case Study} &
\multirow{2}{*}{Computational Cost} &
\multicolumn{5}{c}{Learning Method} \\ 
\cmidrule(lr){3-7}
No. & (minutes) &  HL & VH & EL & SL & PINN\\ \midrule
        & Data Acquisition & 83 & - & - & 142 & - \\
Case 1  & Model Training & 110 & 195 & 600 & 52 & 195 \\
        & Total Cost & 193 & 195 & 600 & 194 & 195\\
\midrule
        & Data Acquisition & 250 & - & - & 363 & - \\
Case 2  & Model Training & 165 & 420 & 840 & 60 & 420 \\
        & Total Cost & 415 & 420 & 840 & 423 & 420\\
\midrule
        & Data Acquisition & 250 & - & - & 363 & - \\
Case 3 & Model Training & 180 & 430 & 880 & 70 & 430 \\
        & Total Cost & 430 & 430 & 880 & 433 & 430\\
\midrule
        & Data Acquisition & 500 & - & - & 625 & - \\
Case 4  & Model Training & 210 & - & - & 85 & 716 \\
        & Total Cost & 710 & - & - & 710 & 716\\
\bottomrule
\end{tabular}
\vspace{-5pt}
\end{table}

\textbf{Network architecture.} For all cases, we will present results obtained using fully-connected networks with 3 hidden layers, each comprising 64 neurons, and with \texttt{tanh}, \texttt{relu}, or \texttt{sin} activation functions. The following experimentations on network architecture were conducted but omitted to keep the paper concise: (1) Experiments on deeper and wider networks did not lead to significant improvement in generalization and safety performance, or qualitative changes to the conclusions we will present; (2) we observe that \texttt{gelu} performs similarly to \texttt{tanh} in terms of the generalization and safety performance.

\textbf{Hardware.} For all case studies, all methods except epigraphical learning are conducted on one workstation with 3.50GHz Xeon E5-1620 v4 CPU and four GeForce GTX 1080 Ti GPU with 11 GB memory. Due to the increased dimensionality of the augmented value in epigraphical learning, we use an A100 GPU with 40 GB memory to achieve convergence. Our empirical results suggest that epigraphical learning is not as data efficient as the hybrid method even with this advantage.

\textbf{Performance metrics.} Since all case studies involve collision avoidance as their state constraints, our analysis will focus on collision rate (Col.\%) as a safety metric. Specifically, collision rate is the probability of sampling an initial state for which closed-loop control of both players using the value network leads to a collision: $Col.\% = N_{pred}/N_{gt}$, where $N_{pred}$ is the number of collision trajectories resulted from the value network and $N_{gt}$ is the number of collision-free trajectories resulted from solving BVPs. Both share the same uniform samples of initial states. Additionally, we report in Case 1 generalization performance of the value networks in terms of their mean absolute approximation errors in value and control inputs along the test state trajectories. The ground truth value and control inputs are derived from BVP.

\textbf{Hypotheses.} The following hypotheses will be tested empirically through the case studies:

(1) \textit{With the same computational budget, HL yields better generalization and safety performance than vanilla PINN, VH, SL, and EL across all presented cases and settings. The key ingredient for high safety performance is the costate loss.}


(2) \textit{The choice of the activation function and its parameters is critical to the safety performance. In general, continuously differentiable activations, e.g., \texttt{tanh} and \texttt{sin}, are better than activations with discontinuous derivatives, e.g., \texttt{relu}.}

\vspace{-0.15in}
\subsection{Case 1: uncontrolled intersection}  
\label{sec:uncontrolled intersection}

\begin{figure}[!ht]
\centering
\vspace{-10pt}
\includegraphics[width=0.96\linewidth]{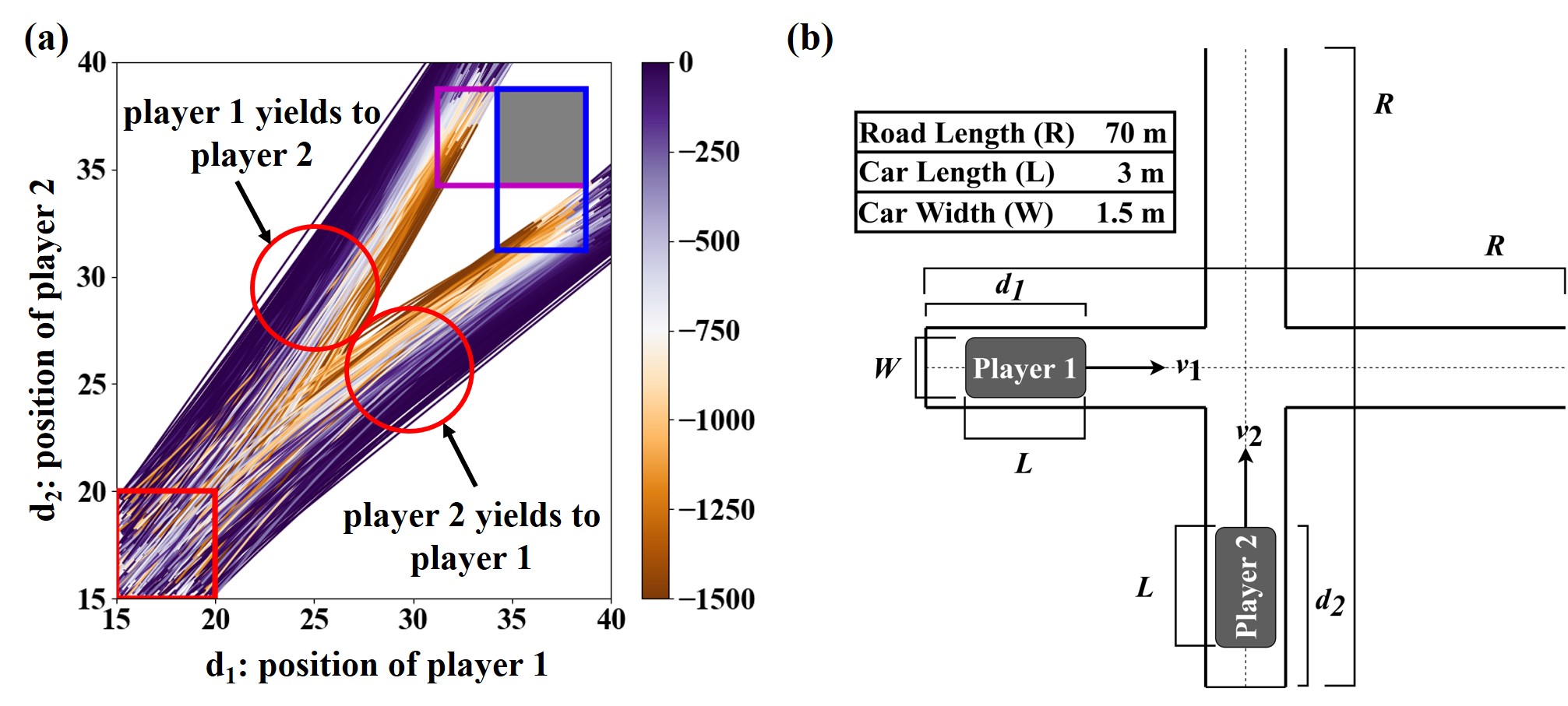}
\caption{(a) State trajectories of players projected to $(d_1, d_2)$. Solid gray box: collision area from the perspective of aggressive players; hollow boxes (magenta for Player 1 and blue for Player 2): collision areas from the perspectives of non-aggressive players. Red box: sampling domain for initial states. Color: Actual values of Player 1. (b) Uncontrolled intersection setup.}
\label{fig:intersection case}
\vspace{-12pt}
\end{figure}

\textbf{Experiment setup.} 
The schematics of the uncontrolled intersection case and the parameters ($R$, $L$, and $W$ for road length, car length, and car width, respectively) are depicted in Fig.~\ref{fig:intersection case}. Each player is represented by two state variables: location ($d_i$) and speed ($v_i$), which together form the state of the player as $x_i := (d_i, v_i)$. The shared dynamics between the players follow the equations $\dot{d}_i = v_i$ and $\dot{v}_i = u_i$, where $u_i \in [-5, 10] m/s^2$ represents the scalar control input, i.e., the acceleration of the player. The instantaneous loss is
\begin{equation}
    l_i(u_i) = u_i^2,
\end{equation}
and the player type-dependent state constraint is
\begin{equation}
    c_i(\textbf{x}_i;\theta) = \delta(d_i,\theta_i)\delta(d_{-i},1)
\end{equation}
Here $\delta(d,\theta) = 1$ iff $d \in [R/2 - \theta W/2, (R+W)/2 + L]$ or otherwise $\delta(d,\theta) = 0$. 
$\theta \in \Theta := \{1, 5\}$ represents the aggressive (\texttt{a}) or non-aggressive (\texttt{na}) type of a player, where the non-aggressive player adopts a larger collision zone, see hollow boxes in Fig.~\ref{fig:intersection case}.
The terminal loss is defined to incentivize players to move across the intersection and restore nominal speed: 
\begin{equation}
g_i(x_i) = -\mu d_i(T) + (v_{i}(T)-\bar{v})^2,
\label{eq:case 1_terminal loss}
\end{equation}
where $\mu = 10^{-6}$, $\bar{v} = 18 m/s$, and $T = 3s$. For hybrid, value-hardening, and vanilla PINN, we treat the state constraint as a penalty in a modified instantaneous loss: 
\begin{equation}
    \tilde{l}_i(\textbf{x}_i, u_i;\theta) = l_i(u_i) + b \sigma(d_i,\theta_i)\sigma(d_{-i},1),
\end{equation}
where 
\begin{equation}
\begin{aligned}
    \sigma(d,\theta) = & \left(1+\exp(-\gamma (d-R/2+\theta W/2))\right)^{-1}\\
    & \left(1+\exp(\gamma (d-(R+W)/2-L))\right)^{-1},
\end{aligned}
\end{equation}
$\gamma = 5$ is a shape parameter and $b = 10^4$ is chosen to be large enough to avoid collisions, and cause a large Lipschitz constant in the resulting value functions. 

\textbf{Data.} 
For supervised learning, 1.7k ground truth trajectories are generated from initial states uniformly sampled in $\mathcal{X}_{GT} := [15, 20]m \times [18, 25]m/s$ by solving Eq.~\eqref{eq:pmp}. Each trajectory consists of 31 $\times$ 2 data points (sampled with a time interval of $0.1s$ and for two players), resulting in a total of 105.4k data points. For vanilla PINN and value hardening, 122k states are sampled uniformly in $\mathcal{X}_{HJ} := [15, 105]m \times [15, 32]m/s$. For hybrid learning, 1k ground truth trajectories (62k data points) are uniformly sampled in $\mathcal{X}_{GT}$, and 60k states are uniformly sampled in $\mathcal{X}_{HJ}$. For epigraphical learning, we first gather a sample of 200k states from $\mathcal{X}_{HJ}$ to ensure adherence to the boundary conditions. 
Subsequently, additional 110k states are sampled from $\mathcal{X}_{HJ}$ every 30k training iterations, resulting in a total of 1300k sampled data points upon completion of the training process.

For the auxiliary state, recall that its initial value represents the player's value of the game. In the intersection case, the best-case loss is $-1.05\times 10^{-4}$ with zero collisions and control efforts, while the worst-case loss without collision is $300$ where the player constantly uses the maximum acceleration or deceleration. Hence we uniformly sample $z_i \in[-1.05\times 10^{-4}, 300]$. The same sampling procedure is applied to all sub-cases with enumeration of player types: (\texttt{a}, \texttt{a}), (\texttt{na}, \texttt{a}), (\texttt{a}, \texttt{na}), and (\texttt{na}, \texttt{na}). 

The selection of state spaces to sample from, namely $\mathcal{X}_{GT}$ and $\mathcal{X}_{HJ}$, is based on various factors: In the case of ground truth trajectories, the initial states for both players are uniformly sampled from identical domains. This is because informative collision and near-collision cases often occur when players start from similar states. Additionally, the range of locations for supervised data is chosen as $[15, 20]m$ to increase the likelihood of sampling informative trajectories within the specified time window. The speed range of $[18, 25]m/s$ is selected based on typical vehicle speed limits. For PINN and variants, the sample space $\mathcal{X}_{HJ}$ approximately covers all states that players can reach within the time window. It is noteworthy that within $\mathcal{X}_{HJ} \times \mathcal{X}_{HJ}$, about 20\% of the states will induce collisions. Adaptive sampling for PINN such as in \cite{jagtap2020adaptive} can potentially improve the data efficiency further but is not studied in this paper. 

\begin{figure}[!ht]
\vspace{-10pt}
\centering
\includegraphics[width=0.96\linewidth]{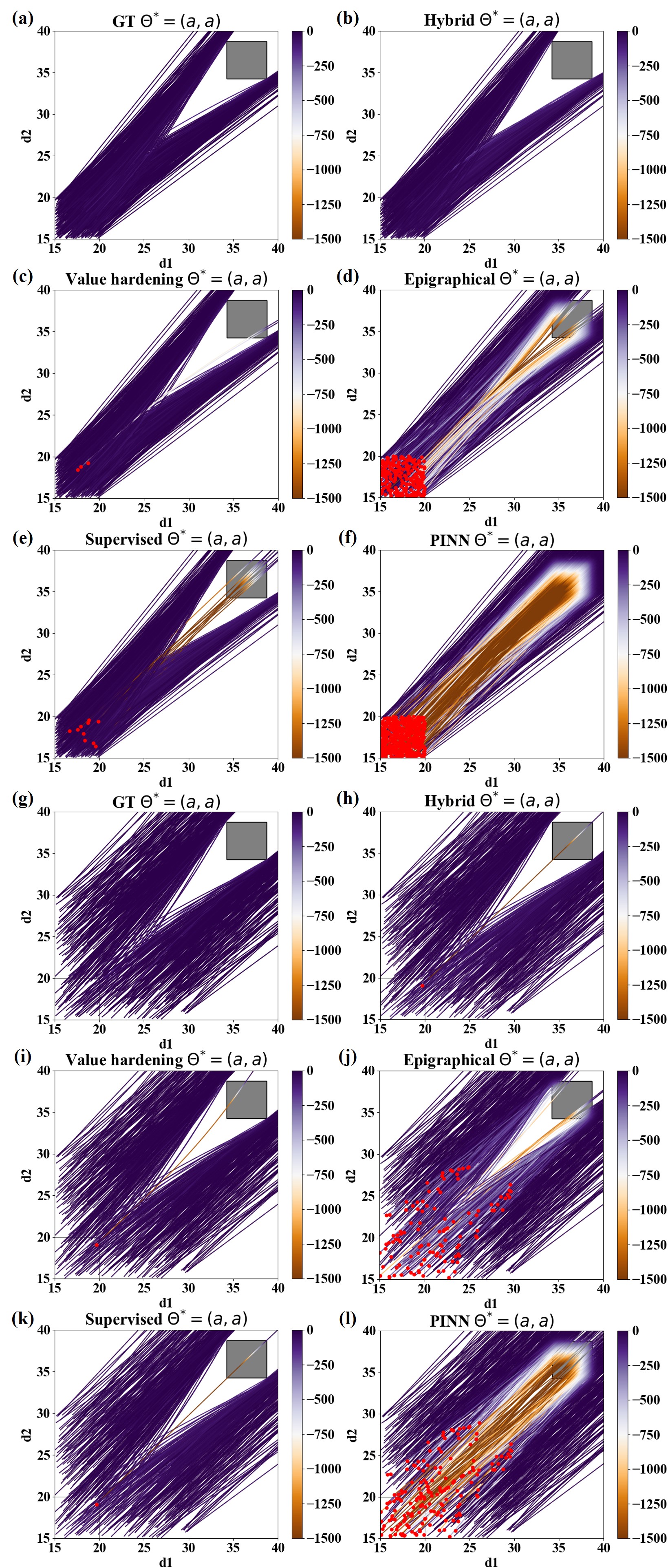}
\caption{(a), (g): Ground truth trajectories (projected to $d_1$-$d_2$) for $\mathcal{X}_{GT}$ and $\mathcal{X}_{XP}$, respectively. (b-f), (h-l): Trajectories generated using hybrid, value hardening, epigraphical, supervised, and vanilla PINN methods under $\mathcal{X}_{GT}$ and $\mathcal{X}_{XP}$, respectively. Color: Actual equilibrial values of Player 1 along the trajectories. Trajectories with inevitable collisions are removed for clearer comparison on safety performance. Red dots represent initial states with \textit{avoidable} collisions.}
\vspace{-10pt}
\label{fig:complete info}
\end{figure}

\textbf{Training.} All training problems except epigraphical learning are solved using the Adam optimizer with a fixed learning rate of $2 \times 10^{-5}$. For supervised learning, the networks are trained for 100k iterations. For vanilla PINN, we adopt the curriculum learning method proposed in~\cite{deepreach}. Specifically, we first train the networks for 10k iterations using 122k uniformly sampled boundary states at the terminal time. We then refine the networks for 260k gradient descent steps, with states sampled from an expanding time window starting from the terminal. For value hardening, we follow the same learning procedure, but we soften the collision penalty using sigmoid functions and gradually increased the shape parameter of the sigmoid to harden the penalty. To keep the computational cost of value hardening similar to that of the hybrid, we use 5.4k training iterations for each hardening step for a total of 50 steps. For the hybrid method, we pretrain the networks for 100k iterations using the supervised data and combine the supervised data with states sampled from an expanding time window starting from the terminal time to minimize $L_1 + L_2$ for 100k iterations. For epigraphical learning, we first train the network to fulfill the boundary condition over 50k iterations. Subsequently, we refine the network through 3k gradient steps per epoch, encompassing a total of 10 epochs for every 30k training iterations. The network refinement process spans 300k training iterations in total.

\begin{table}[!ht]
\vspace{-10pt}
\scriptsize
\centering
\caption{Generalization and safety performance (collision rate) on complete-information games. HL, VH, EL, SL, PINN are for hybrid, value hardening, epigraphical, supervised, and vanilla physics-informed neural network methods, respectively.}
\label{table:complete info}
\begin{tabular}{lccccc}
\toprule
\multirow{2}{*}{Test} &
\multirow{2}{*}{Player} &
\multirow{2}{*}{Learning} &
\multicolumn{3}{c}{Metrics} \\ 
\cmidrule(lr){4-6}
Domain & Types & Method & $|\vartheta -\hat\vartheta|$ $\downarrow$	
  & $|u - \hat u|$ $\downarrow$ & Col.\% $\downarrow$\\ \midrule
  &  & HL &  \textbf{0.46} & \textbf{0.09 $\pm$ 0.10} & \textbf{0.00\%} \\
  &  &  VH &  4.17 & 0.34 $\pm$ 0.19 & 0.67\% \\
  & (\texttt{a}, \texttt{a}) & EL &  28.30 & 0.85 $\pm$ 3.92 & 42.3\% \\
  &  & SL & 0.57 &   0.12 $\pm$ 0.36 & 1.67\% \\
  &  & PINN & 3.39  & 0.96 $\pm$ 4.19 & 84.8\%\\
  \cmidrule(lr){3-6}
  &  & HL & \textbf{9.43} &  \textbf{0.49 $\pm$ 3.55}  & 3.50\% \\
  &  & VH &  79.35 & 1.10 $\pm$ 5.42 & \textbf{0.50\%} \\ 
  $\mathcal{X}_{GT}$ & (\texttt{a}, \texttt{na}) & EL & 123.79 & 2.24 $\pm$ 20.8 & 42.7\% \\
  &  & SL & 10.58 & 0.54 $\pm$ 3.92 & 4.50\% \\
  &  & PINN & 15.33 & 1.27 $\pm$ 7.16 & 83.3\% \\
  \cmidrule(lr){3-6}
  &  & HL & \textbf{1.00} & \textbf{0.04 $\pm$ 0.03}  & \textbf{1.33\%} \\ 
  &  & VH &  21.76 & 0.34 $\pm$ 1.33 & 8.50\%\\
  & (\texttt{na}, \texttt{na}) & EL & 130.53 & 0.66 $\pm$ 5.66 & 16.5\% 
  \\ 
  &  & SL & 3.49  & 0.10 $\pm$ 0.46 & 4.33\%\\
  &  & PINN & 114.67  & 1.88 $\pm$ 13.72 & 83.5\% \\
\midrule
  &  &  HL & \textbf{0.41}  & \textbf{0.09 $\pm$ 0.08}  & \textbf{0.20\%} \\ 
  &  &  VH &  2.03 & 0.20 $\pm$ 0.07 & \textbf{0.20\%}\\ 
  & (\texttt{a}, \texttt{a}) & EL &  11.93 & 0.34 $\pm$ 1.62 & 19.0\% \\
  &  & SL & 0.69 & 0.17 $\pm$ 0.28 & \textbf{0.20\%} \\
  &  & PINN & 1.54 & 0.37 $\pm$ 1.88  & 35.2\% \\
  \cmidrule(lr){3-6}
  &  & HL & \textbf{17.39}  & \textbf{0.46 $\pm$ 3.17} & \textbf{0.10\%} \\ 
  &  & VH &  32.64 & 0.57 $\pm$ 2.71 & 0.20\%\\ 
  $\mathcal{X}_{XP}$ & (\texttt{a}, \texttt{na}) & EL &  62.62 & 0.96 $\pm$ 8.64 & 10.5\% \\
  &  & SL & 19.01 & 0.56 $\pm$ 3.09 & 0.60\% \\
  &  & PINN & 19.57 & 0.58 $\pm$ 3.89 & 31.3\% \\
  \cmidrule(lr){3-6}
  &  & HL & \textbf{1.80} & \textbf{0.10 $\pm$ 0.12} & \textbf{0.00\%}\\
  &  & VH &  11.54 & 0.24 $\pm$ 0.68 & 6.40\% \\ 
  & (\texttt{na}, \texttt{na}) & EL & 63.73 & 0.41 $\pm$ 3.02 & 2.33\% \\ 
  &  & SL & 4.25 & 0.30 $\pm$ 0.72 & 2.20\% \\
  &  & PINN & 60.39 & 0.95 $\pm$ 7.31 & 36.0\%  \\
\bottomrule
\end{tabular}
\vspace{-10pt}
\end{table}

It should be noted that our initial experiment with epigraphical learning led to poor generalization and safety performance. In the results we will present, adaptive activations~\cite{jagtap2020adaptive} and adaptive learning rates are implemented, in addition to the use of a larger computational budget, to slightly improve the performance, which still falls short of that of hybrid learning. 
\begin{figure*}[!ht]
\centering
\includegraphics[width=0.95\linewidth]{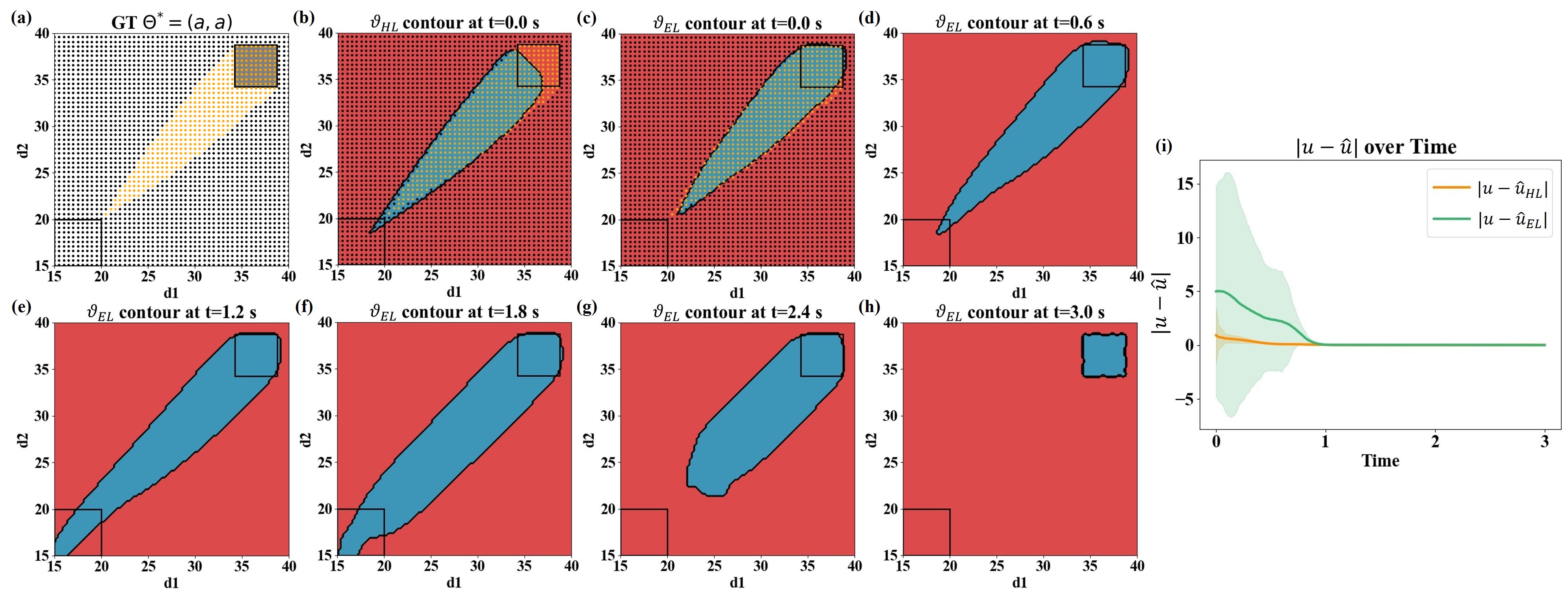}
\caption{(a) Ground truth safe/unsafe initial states projected to $d_1$-$d_2$ frame, where black dots represent collision-free trajectories while orange dots depict trajectories with collision. (b) Value contours at initial time to classify safe/unsafe zones using HL. (c-h): Value contours along time using EL. Blue (red) regions represent unsafe (safe) states. 
(i) Comparison of mean and standard deviation of $|u-\hat u|$ from HL and EL across test trajectories sampled from $\mathcal{X}_{GT}$.}
\vspace{-20pt}
\label{fig:safety_zone}
\end{figure*}

\vspace{-0.15in}
\subsubsection{Results for complete-information games} We generate a separate set of 600 ground truth trajectories for each of the four player type configurations by solving BVPs, with initial states uniformly sampled from $\mathcal{X}_{GT}$. To evaluate generalization performance, we measure the mean absolute errors (MAEs) of value and control input predictions, denoted by $|\vartheta - \hat{\vartheta}|$ and $|u - \hat{u}|$, respectively, across the test trajectories. 
For safety performance, we use the learned value networks to compute the players' closed-loop control inputs and the state trajectories. From all resulting trajectories computed based on test initial states, we report the percentage of collisions that are \textit{avoidable} according to BVP solutions. The performance results are summarized in Tab.~\ref{table:complete info}, where we averaged the performance of (\texttt{a}, \texttt{na}) and (\texttt{na}, \texttt{a}) due to their symmetry. Sample trajectories for (\texttt{a}, \texttt{a}) are shown in Fig.~\ref{fig:complete info}.

To further evaluate the out-of-distribution performance of supervised and hybrid learning, we repeat the tests using 500 uniformly sampled initial states in $\mathcal{X}_{XP}:= [15, 30]m \times [18, 25]m/s$. The results are summarized in the same table and figure. In both tests, the hybrid method demonstrates the best generalization and safety performance. Notably, the vanilla PINN exhibits poor generalization due to value discontinuity. Epigraphical learning performs only better than vanilla PINN in regard of safety. Further inspection shows that epigraphical learning can actually identify the backward reachable sets (i.e., unsafe zones) well, see Fig.~\ref{fig:safety_zone}. To elaborate, given $t \in [0,T]$ and a value network $\hat{V}$ trained for fix player types (\texttt{a}, \texttt{a}), the unsafe zone is defined as $\{\textbf{x} \in \mathcal{X}_{XP} | \hat{V}(\textbf{x}, t) > 0\}$. We approximate the ground truth unsafe zone by computing trajectories of sample initial states in $\mathcal{X}_{XP}$ by solving Eq.~\eqref{eq:pmp} (Fig.~\ref{fig:safety_zone}a). We compare the ground truth with the approximations from hybrid and epigraphical learning in Fig.~\ref{fig:safety_zone}b,c. The results here reveal an important limitation of values approximated through PINN: High empirical accuracy in characterizing the unsafe zone does not necessarily imply high safety performance, such as in the case of epigraphical learning. This is potentially because feedback control requires accurate approximation of the value \textit{gradients} instead of the segmentation of value in space-time (see $|u-\hat u|$ comparison in Fig.~\ref{fig:safety_zone}i). For the same reason, high safety performance does not imply high accuracy in characterizing the unsafe zone either, such as in the case of hybrid learning. We further verify that adding the supervised costate loss to EL improves its safety performance to be comparable with that of HL. See Sec.~\ref{sec:EL explain} for details.

\begin{figure}[!ht]
\vspace{-10pt}
\centering
\includegraphics[width=0.96\linewidth]{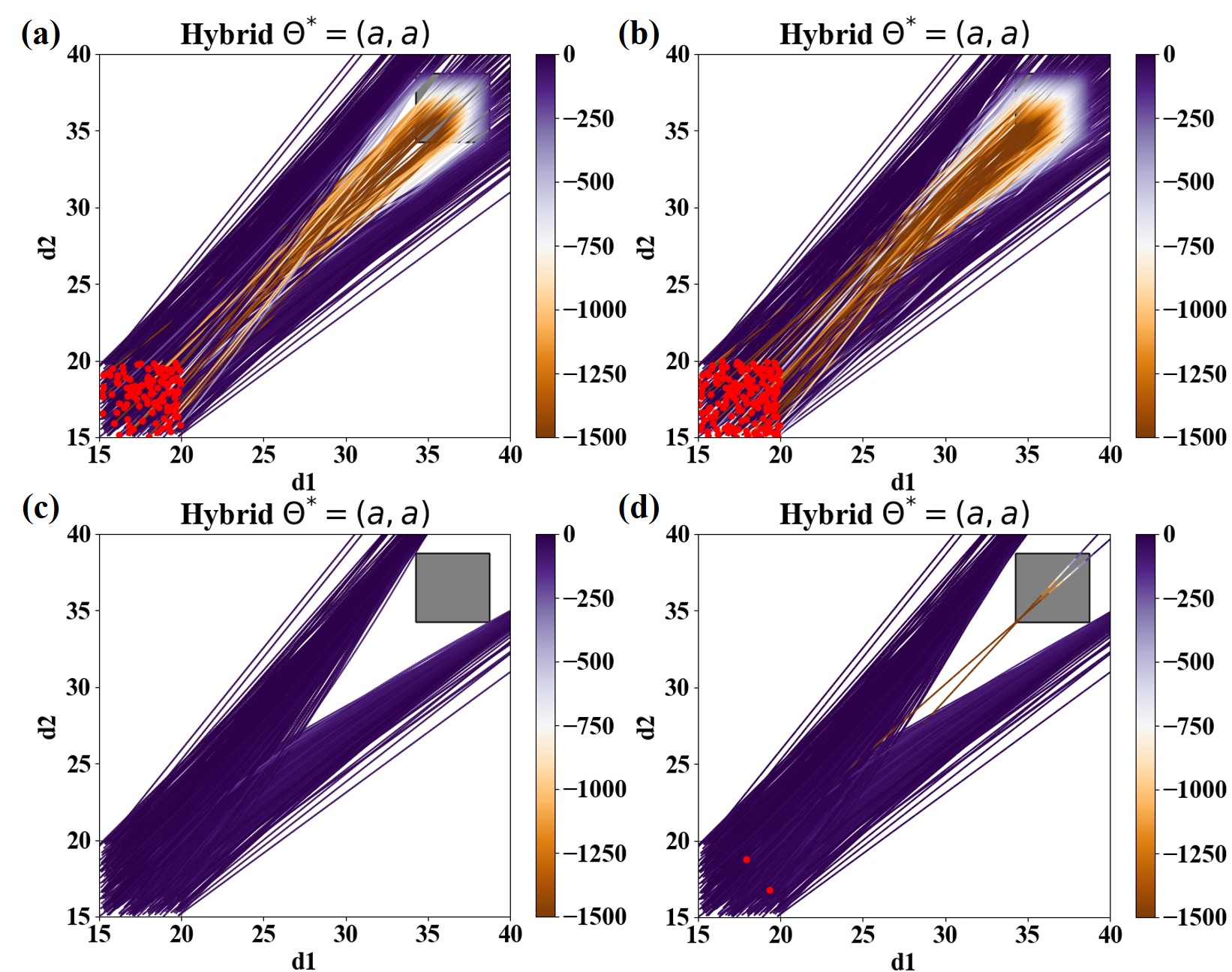}
\caption{Trajectories generated using neural networks with (a) \texttt{relu} and (b) \texttt{sin} activation functions and using $L^1$ for boundary norm (for \texttt{tanh}, refer to Fig.~\ref{fig:complete info}); trajectories generated using (c) $L^1$- and (d) $L^{2}$-norms for the boundary values and using $\texttt{tanh}$ for activation. All trajectories are based on hybrid learning.}
\vspace{-10pt}
\label{fig:ablation}
\end{figure}

\textbf{Ablation studies.} We conduct ablation studies to understand the effects of activation functions and the norm of the boundary loss on model performance. Safety results are summarized in Tab.~\ref{table:activation} for player types (\texttt{a}, \texttt{a}) and using the hybrid learning method, with training and testing conducted in $\mathcal{X}_{GT}$. The corresponding trajectories are visualized in Fig.~\ref{fig:ablation}. The results indicate that (1) the choice of the activation function significantly affects the resultant models, with \texttt{tanh} outperforming \texttt{relu} and \texttt{sin}, and (2) the choice of the boundary norm does not have a significant influence.

\textit{Remarks}:
We note that \texttt{relu} networks have been shown to converge to piecewise smooth functions in a supervised setting~\cite{petersen2018optimal}. However, convergence in the PINN setting requires continuity of the network and its gradient~\cite{shin2020convergence}, which \texttt{relu} does not offer. Our results are consistent with those of \cite{jagtap2020adaptive}, where \texttt{relu} underperforms in solving PDEs. We note, however, that smooth variants of \texttt{relu} such as \texttt{gelu} can achieve performance comparable to that of \texttt{tanh}. We also note that while \texttt{sin} does not perform well for Case 1, it achieves comparable performance to \texttt{tanh} in Cases 2-4 (see Sec.~\ref{sec:narrow-road},~\ref{sec:lane-change} and~\ref{sec:drone-avoidance}). This result suggests that fine-tuning of the frequency parameter of \texttt{sin} is necessary and case-dependent~\cite{siren}.

\begin{table}[!ht]
\vspace{-5pt}
\centering
    \caption{Safety performance (collision rate) w/ different activation functions (w/ $L^1$) and boundary norms (w/ \texttt{tanh})}
    \label{table:activation}
\begin{tabular}{lccccc}
\toprule
\multirow{2}{*}{Method} & \multicolumn{3}{c}{Activation } & \multicolumn{2}{c}{Boundary Norm} \\ \cmidrule(lr){2-4}\cmidrule(lr){5-6} 
& \texttt{tanh}   & \texttt{relu}  & \texttt{sin}  & $L^1$   & $L^2$  \\ \midrule
Hybrid  & \textbf{0.00\%} & 19.8\%   & 28.7\%   & \textbf{0.00\%}  & 0.4\%  \\ 
Value hardening  & 0.67\% & 85.1\%   & 84.6\%   & -  & -  \\ 
Epigraphical  & 42.3\% & 78.8\%   & 89.8\%   & -  & -  \\
Supervised  & 1.67\%  & \textbf{2.50\%} & \textbf{19.5\%} & - & - \\
Physics-informed  & 84.8\% & 84.0\%  & 84.7\% & -  & -   \\
\bottomrule
\end{tabular}
\vspace{-5pt}
\end{table}


\subsubsection{Results for incomplete-information games} In games with incomplete information, we investigate the effectiveness of using value networks both for closed-loop control and for belief updates: Each player is uncertain about the types of the other players and therefore holds a belief about their fellow player's types. A belief is a probability distribution over the type space and is updated over time as the player observes new actions from their fellow player. We examine two belief update settings: the first assumes that players have common prior belief and synchronized belief dynamics~\cite{harsanyi1967games}. In other words, Player $i$ knows about Player $j$'s uncertainty about Player $i$'s type. We refer to players in this setting as ``empathetic''. The second setting is non-empathetic, where Player $i$ falsely assumes that Player $j$ has full knowledge about Player $i$'s type. We follow \cite{chen2021shall} to simulate the state and belief dynamics: We model Player $i$ to continuously update its belief based on observations, and then determine its next control inputs based on the value network parameterized by the most likely type of Player $j$ as well as Player $i$'s truth type. We evaluate the efficacy of the hybrid and supervised methods, which achieve the best performance across cases, by measuring their safety performance in incomplete-information settings. The simulations use the same initial states as tests in the complete-information games. 

\textbf{Empathetic belief update.} 
We consider the case where players can take one of the two types: $\Theta = \{a, na\}$.
Let $\mathcal{D}_t=\{({\textbf{x}}(k), {\textbf{u}}(k))\}_{k=1}^t$ be a finite set of observed states and control inputs of both players accumulated up to time $t$. 
Let $p_i(t):= \Pr(\theta_i = a ~|~ \mathcal{D}_{t-1})$ be the belief of Player $j$ about Player $i$ at the beginning of time step $t$, and $q_i^{\hat{\boldsymbol{\theta}}}(t) := \Pr(u_i(t)~|~\textbf{x}(t),\hat{\boldsymbol{\theta}})$ where $\hat{\boldsymbol{\theta}} \in \{(a,a), (a,na), (na, a), (na, na)\}$ is a point estimate of $\boldsymbol{\theta}$ based on the current beliefs $\textbf{p}$.

We assume Player $i$'s control policy follows a Boltzmann distribution:
\begin{equation}
    q_i^{\hat{\boldsymbol{\theta}}}(t) = \frac{e^{ h_i(\textbf{x}_i(t),u_i(t), t;\hat{\boldsymbol{\theta}})}}{\sum_{\mathcal{U}} e^{ h_i(\textbf{x}_i(t),u_i',t;\hat{\boldsymbol{\theta}})}},
    \label{eq:decision}
\end{equation}
where
\begin{equation}
    h_i(\textbf{x}_i(t),u_i(t), t;\hat{\boldsymbol{\theta}}) = \nabla_{\textbf{x}_i}  \textbf{f}_i^T \vartheta_i^{\hat{\boldsymbol{\theta}}} - \tilde{l}_i^{\hat{\theta}_i}.
\end{equation}
$\vartheta_i^{\hat{\boldsymbol{\theta}}}$ is Player $i$'s approximated value if the game is played with player types $\hat{\boldsymbol{\theta}}$, and $\tilde{l}_i^{\hat{\theta}_i}$ is the instantaneous loss that incorporates the collision penalty if Player $i$ is of type $\hat{\theta}_i$.  

Denote the marginal by $q_i^{\hat{\theta}_i}(t):= \Pr(u_i(t)|\textbf{x}(t), \hat{\theta}_i)$, we have
\begin{equation}
    q_i^{\hat{\theta}_i}(t) = q_i^{(\hat{\theta}_i,a)}(t)p_{-i}(t) + q_i^{(\hat{\theta}_i,na)}(t)(1-p_{-i}(t)).
\end{equation}
Given the observations $\mathcal{D}_t$, $p_i$ follows a Bayes update:
\begin{equation}
    p_i(t+1) = \frac{q_i^{a}(t) p_i(t)}{q_i^{a}(t) p_i(t) + q_i^{na}(t)(1-p_i(t))}.
    \label{eq:beliefupdate}
\end{equation}

\textit{Remarks}:
(1) If any element of $\textbf{p}(t)$ is mistakenly assigned a zero probability, this mistake cannot be corrected in future updates. To address this, we modify $\textbf{p}(t)$ using
\begin{equation}
    \textbf{p}(t) \Leftarrow (1-\epsilon)\textbf{p}(t) + \epsilon \textbf{p}(0),
    \label{eq:preprocess}
\end{equation}
before its next update and set the learning rate $1-\epsilon = 0.95$. $\textbf{p}(0)$ represents the initial belief. 
(2) To make Eq.~\eqref{eq:decision} more tractable, we discretize the space of control inputs as $\mathcal{U}:=\{-5, -4, ..., 0, ..., 10\}m/s^2$. Additionally, we used discrete time steps with a time interval of 0.05 seconds to simulate the interactions. 
(3) We test two settings of initial beliefs. In the first setting, each player believes that the other player has a probability of 80\% of being aggressive; in the second setting, the probability is 20\%. These initial beliefs correspond to $\textbf{p}(0) = (0.8, 0.2)$ and $\textbf{p}(0) = (0.2, 0.8)$, respectively. While a more extensive test over the initial belief space could be interesting, it is beyond the scope of this study.

\textbf{Non-empathetic belief update.} A non-empathetic player updates its belief about the other player's type by assuming that his type is known. Let the true types be $\boldsymbol{\theta}^*$. Player $-i$'s belief about Player $i$'s type now becomes a conditional $p'_i(t):= \Pr(\theta_i = a | \mathcal{D}_{t-1}, \theta_{-i}^*)$.
The Bayes update of $p'_i(t)$ follows:
\begin{equation}
    p'_i(t+1) = \frac{q_i^{(a,\theta_{-i}^*)}(t) p'_i(t)}{q_i^{(a, \theta_{-i}^*)}(t)p'_i(t) + q_i^{(na, \theta_{-i}^*)}(t)(1-p'_i(t)) }.
    \label{eq:beliefupdate_ne}
\end{equation}
Consequently, each player starts with its own belief, which are not necessarily common during the interaction. 

\textbf{Control policy.} Given the beliefs $p_i(t)$ or $p'_i(t)$, Player $-i$ finds the most likely type of Player $i$. The control policy of Player $i$ is determined by the value function corresponding to $(\theta^*_i, \hat{\theta}_{-i})$. It is worth noting that Player $i$ employs a policy that is consistent with its true type, even if Player $j$ holds an incorrect belief about Player $i$, which Player $i$ acknowledges in the empathetic setting. This setup allows players to signal their own types through their actions.

\textbf{Simulation results.} We present simulated interactions between two players at an uncontrolled intersection in an incomplete-information setting. The simulations are performed on a grid that enumerates the following settings: (empathetic, non-empathetic) $\times$ (correct prior, wrong prior) $\times$ (aggressive, non-aggressive), where both players have identical settings to limit the scope. For each setting, we evaluate the safety performance of the value approximation models learned through the hybrid and supervised methods using test samples from $\mathcal{X}_{GT}$. 
Tab.~\ref{table:inference} summarizes the results that the hybrid models have a lower chance of collision than the supervised ones under all settings. 

\begin{table}[h!]
\vspace{-5pt}
\centering
    \caption{Collision rate in uncontrolled intersections with incomplete information: e - empathetic, ne - non-empathetic}
    \label{table:inference}
    \begin{tabular}{lcccc} 
      \toprule
      Belief update model & Initial belief & True type & Hybrid & Supervised\\\midrule
      (e,e) & (a,a) & (a,a) & \textbf{0.00}\% & \textbf{0.00}\%\\
      (ne,ne) & (a,a) & (a,a) & \textbf{0.00}\% & \textbf{0.00}\%\\
      (e,e) & (na,na) & (na,na) & \textbf{0.67}\% & 6.67\%\\
      (ne,ne) & (na,na) & (na,na) & \textbf{0.67}\% & 6.67\%\\
      (e,e) & (a,a) & (na,na) & \textbf{2.00}\% & 2.67\%\\
      (ne,ne) & (a,a) & (na,na) & \textbf{2.67}\% & \textbf{2.67}\%\\
      (e,e) & (na,na) & (a,a) & \textbf{2.00}\% & 8.00\%\\
      (ne,ne) & (na,na) & (a,a) & \textbf{2.67}\% & 4.00\%\\
      \bottomrule
    \end{tabular}
    \captionsetup{font={footnotesize}}
\vspace{-10pt}
\end{table}

\vspace{-0.2in}
\subsection{Case 2: narrow road collision avoidance} 
\label{sec:narrow-road}
\begin{figure}[!h]
\vspace{-5pt}
\centering
\includegraphics[width=0.87\linewidth]{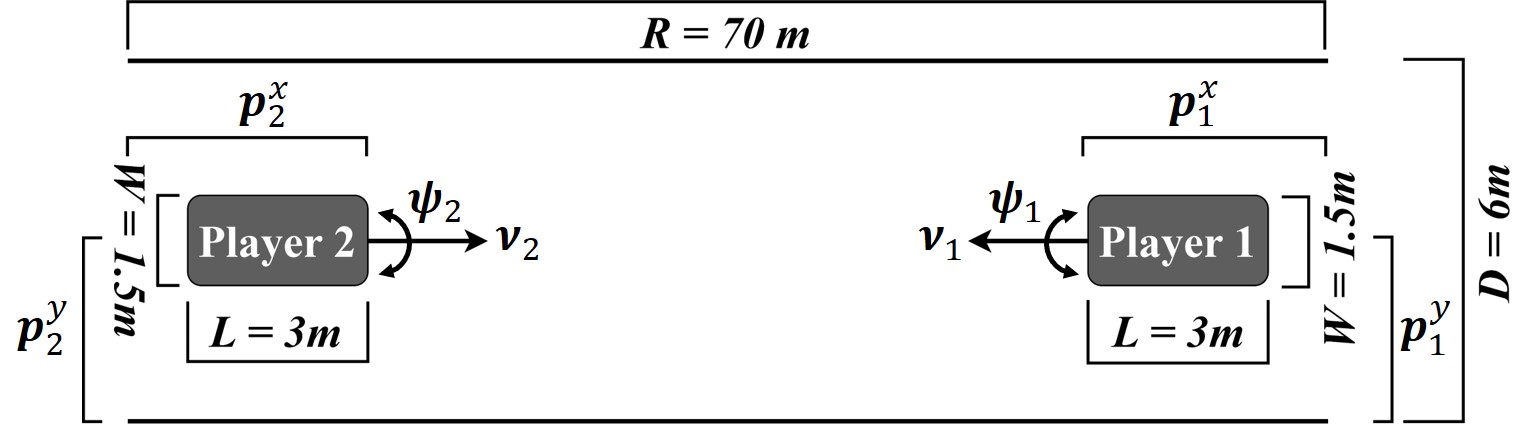}
\caption{Narrow road collision avoidance setup with two players.}
\label{fig:narrow road case}
\vspace{-5pt}
\end{figure}


\textbf{Experiment setup.} 
The schematic is depicted in Fig.~\ref{fig:narrow road case}, where the states of Player $i$ consist of its location ($p^{x}_i$, $p^{y}_i$), orientation ($\psi_i$), and speed ($v_i$), denoted as $x_i := [p^{x}_i, p^{y}_i, \psi_i, v_i]^T$. The system dynamics is modeled using a unicycle model: 
\begin{eqnarray}
\begin{aligned}
\left[
\begin{array}{c}
    \dot p^{x}_i \\
    \dot p^{y}_i \\
    \dot \psi_i \\
    \dot v_i \\
\end{array}
\right]
=
\left[
\begin{array}{c}
    v_i\cos(\psi_i) \\
    v_i\sin(\psi_i) \\
    \omega_i \\
    u_i \\
\end{array}
\right],
\end{aligned}
\end{eqnarray} 
where $\omega_i \in [-1, 1] rad/s$ and $u_i \in [-5, 10] m/s^2$ are control inputs that represent angular velocity and acceleration, respectively. The instantaneous loss incorporates control effort:
\begin{equation}
\begin{aligned}
    & l_i(x_i, u_i;\theta_i) = k\omega_i^2 + u_i^2,
\end{aligned}
\end{equation}
where $k=100$. The state constraint is:
\begin{equation}
\begin{aligned}
    c_i(\textbf{x}_i) = \eta - \sqrt{((R - p^{x}_2) - p^{x}_1)^2 + (p^{y}_2 - p^{y}_1)^2}.
\end{aligned}
\label{eq:case 2_state constraint}
\end{equation}
where $\eta = 1.5 m$ and $R = 70 m$. $c_i(\cdot) > 0$ is considered as a collision incident. The parameter $R$ represents the length of the road, 
and $\eta=1.5 m$ is the collision threshold. The terminal loss is designed to encourage players to move along the lane and restore nominal speed:
\begin{equation}
g_i(x_i) = -\mu p^x_{i}(T) + (v_{i}(T)-\bar{v})^2 + (p^y_{i}(T)-\bar{p}^{y})^2,
\label{eq:case 2_terminal loss}
\end{equation}
where $\mu = 10^{-6}$, $\bar{v} = 18 m/s$, $\bar{p}^{y} = 3 m$, and $T = 3s$.
For hybrid, value-hardening, and vanilla PINN, we treat the state constraint as a penalty in a modified instantaneous loss:
\begin{equation}
\begin{aligned}
    & \tilde{l}_i(\textbf{x}_i, \omega_i, u_i) = k\omega_i^2 + u_i^2 + b\sigma(\textbf{x}_i,\eta),
\end{aligned}
\end{equation}
where the penalty function is defined as 
\begin{equation}
\begin{aligned}
    \sigma(\textbf{x}_i,\eta) = \left(1+\exp(-\gamma c_i(\textbf{x}_i)\right))^{-1}. \notag
\end{aligned}
\end{equation}
The parameter $b$ is set to $10^4$ to impose a high penalty on collision, while $\gamma=5$ is a shape parameter.

\begin{figure}[!ht]
\centering
\includegraphics[width=0.83\linewidth]{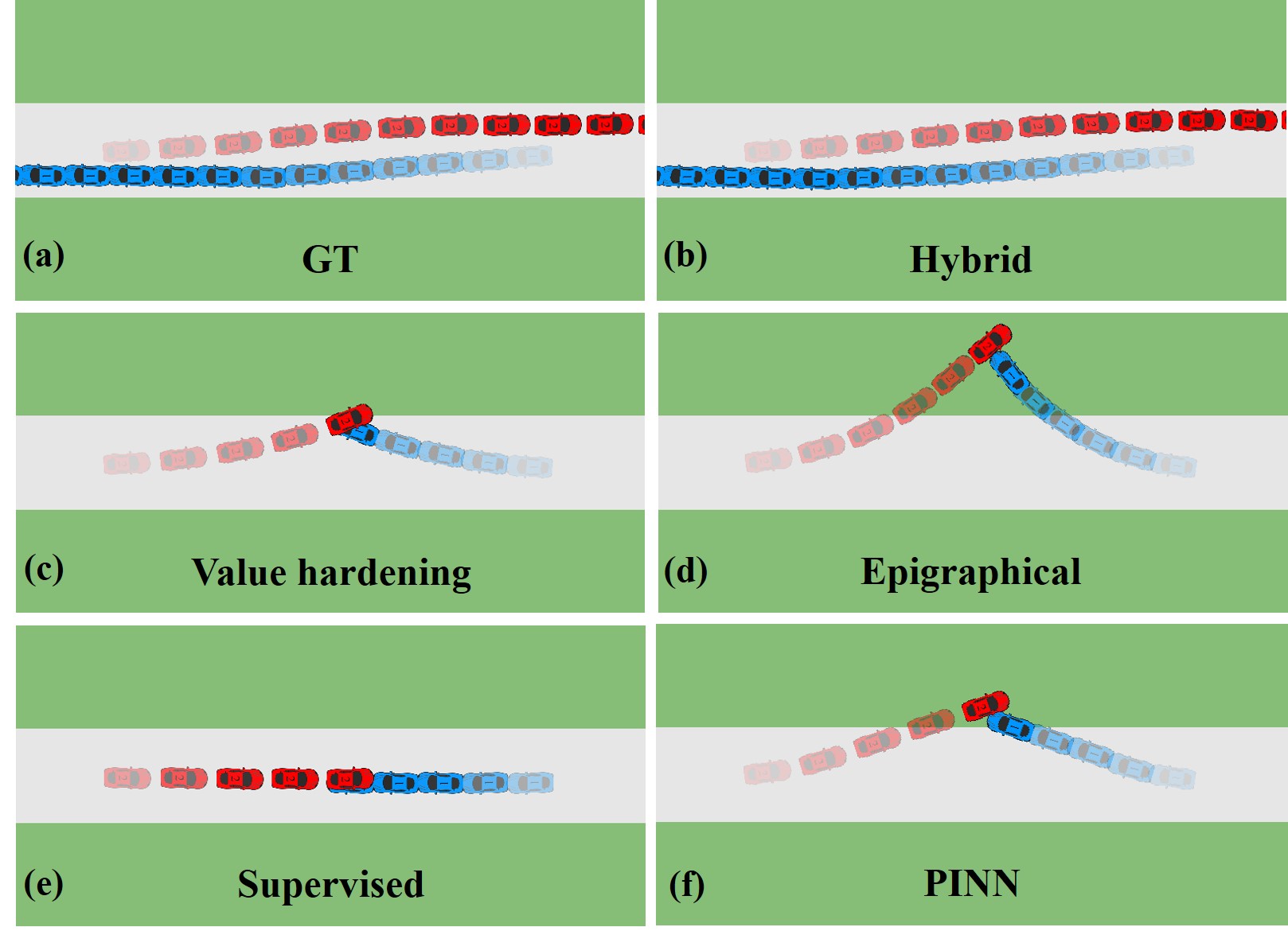}
\caption{Narrow road collision avoidance visualization: (a): Ground truth safe trajectory. Transparency reduces along time. (b-e): Trajectories generated using hybrid, value hardening, epigraphical, supervised, and vanilla PINN models, respectively.}
\label{fig:aviodance_visualization}
\vspace{-15pt}
\end{figure}

\begin{figure}[!ht]
\vspace{-0pt}
\centering
\includegraphics[width=0.96\linewidth]{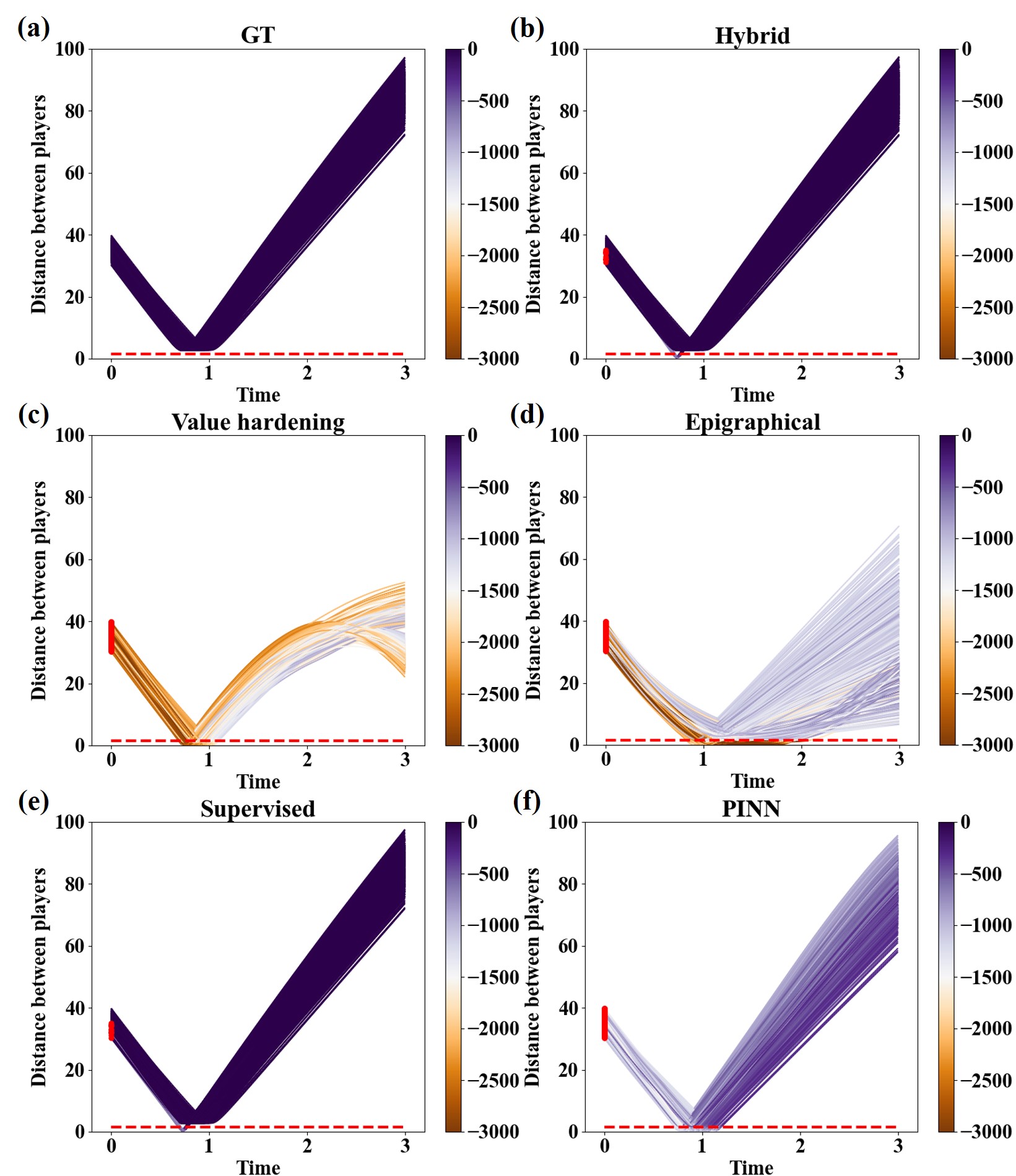}
\caption{(a): Ground truth distance between players over time for $\mathcal{X}_{GT}$. (b-e): Distance between players over time using hybrid, value hardening, epigraphical, supervised, and vanilla PINN under $\mathcal{X}_{GT}$, respectively. Red dashed line represents the threshold distance for collision.}
\label{fig:aviodance_case}
\vspace{-21pt}
\end{figure}

\textbf{Data.} 
For supervised learning, we generate 1.45k ground truth trajectories by uniformly sampling initial states from $\mathcal{X}_{GT}:= [15, 20]m \times [2.25, 3.75]m \times [-\pi/180, \pi/180] rad \times [18, 25]m/s$, resulting in a total of 89.9k data points. 
For vanilla PINN and its value-hardening variant, we uniformly sample 122k states from $\mathcal{X}_{HJ}:= [15, 90]m \times [0, 6]m \times [-0.15, 0.18]rad \times [18, 25]m/s$.
For hybrid learning, we generate 1k ground truth trajectories (62k data points) by uniformly sampling initial states from $[15, 20]m \times [2.25, 3.75]m \times [-\pi/180, \pi/180] rad \times [18, 25]m/s$ and sample 60k states uniformly from $[15, 90]m \times [0, 6]m \times [-0.15, 0.18]rad \times [18, 25]m/s$. For epigraphical learning, we introduce an auxiliary state $z_i$ with a range of $[-9\times 10^{-5}, 300]$ to account for both the best- and worst-case scenarios. We employ the same settings as in Case 1 for the remaining aspects of the experiment.

\textbf{Training.} For vanilla PINN, we pretrain the networks for 10k iterations using 122k uniformly sampled boundary states and then train them for 430k iterations. For value hardening, we use 8.8k training iterations for each hardening step and a total of 50 steps for a fair comparison. The remaining settings are the same as those in Case 1.

\begin{table}[!ht]
\vspace{-3pt}
\centering
    \caption{Collision rate w/ different activation functions
    }
    \label{table:avoidance_safety}
\begin{tabular}{lcccccc}
\toprule
\multirow{2}{*}{Test} &
\multirow{2}{*}{Activation} & & \multicolumn{3}{c}{Learning Method}  \\ \cmidrule(lr){3-7}  
Domain & Functions
& \textbf{HL} & \textbf{VH} & \textbf{EL} & \textbf{SL} & \textbf{PINN} \\ \midrule
& \texttt{tanh}  & \textbf{1.67}\%  & 95.2\% & 48.3\% & 2.17\% & 81.3\% \\
$\mathcal{X}_{GT}$ & \texttt{relu}  & \textbf{65.2\%} & 98.2\% & 70.5\% & 67.7\% & 83.8\% \\
& \texttt{sine}  & \textbf{1.67}\% & 98.5\% & 69.7\% & 3.17\%  & 98.3\% \\ 
\bottomrule
\end{tabular}
\vspace{-5pt}
\end{table}

\textbf{Results.} We evaluate the safety performance of the methods on a test set of 600 ground truth collision-free trajectories with initial states drawn from $\mathcal{X}_{GT}$. The results are summarized in Tab.~\ref{table:avoidance_safety}, and the distance between players during interactions is visualized in Fig.\ref{fig:aviodance_case}. Similar to Case 1, the hybrid learning method outperformed the others. 

We notice that value hardening fails to generalize well in this higher-dimensional case (and in Case 3). We hypothesize that vanilla PINN, which value hardening is based on, is less scalable in compute than hybrid PINN as the state dimensionality increases.     

\begin{figure}[!ht]
\vspace{-10pt}
\centering
\includegraphics[width=1\linewidth]{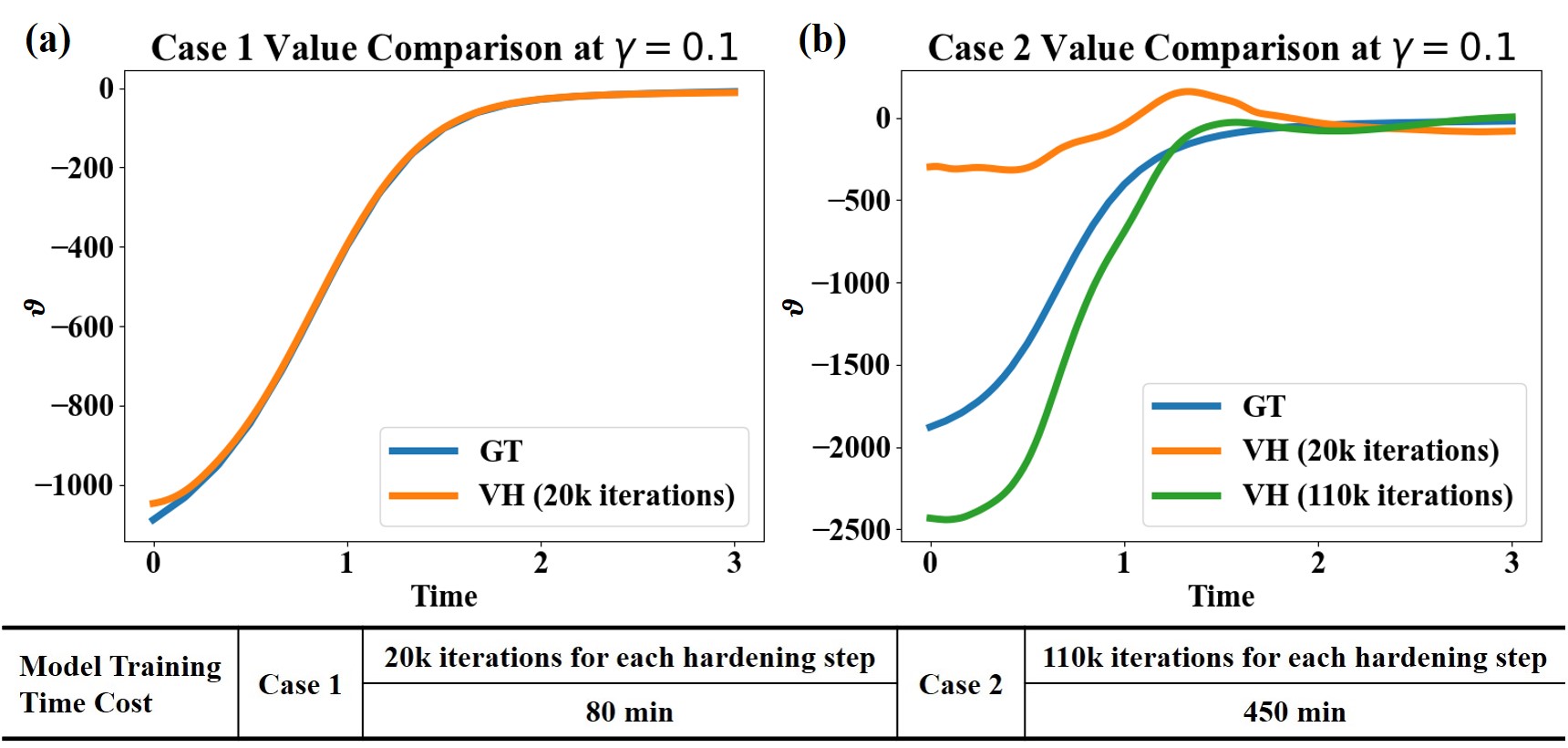}\\
\caption{(a): Value hardening uses 20k training iterations for each hardening step, for a total of 10 steps to converge to ground truth in Case 1. (b) Value hardening uses 20k/110k training iterations for each hardening step, for a total of 10 steps to converge to ground truth in Case 2. Compared to Case 1, value hardening takes around 5.6 times longer to converge to the ground truth in Case 2.}
\label{fig:value_comparison}
\vspace{-12pt}
\end{figure}

While the relationship between learning dynamics of PINN and state dimensionality is yet to be understood, here we empirically show that value hardening PINN requires significantly higher compute to converge in Case 2 due to its higher state dimensionality. To make this empirical study more tractable, we use a mildly softened collision penalty with $\gamma = 0.1$ in both Case 1 and 2. We uniformly sample 122k states from $X_{HJ}$ and train the model using 10 hardening steps until $\gamma$ reaches 0.1. To visualize the convergence, in Figure \ref{fig:value_comparison} we show the value along a randomly chosen equilibrium trajectory derived from PMP for Case 1 (left) and Case 2 (right). We can see that by 20k iterations, value hardening already converges to the ground truth in Case 1, while in Case 2, convergence requires more than 110k iterations.  


\vspace{-0.1in}
\subsection{Case 3: double-lane change}
\label{sec:lane-change}

\begin{figure}[!ht]
\vspace{-5pt}
\centering
\includegraphics[width=0.87\linewidth]{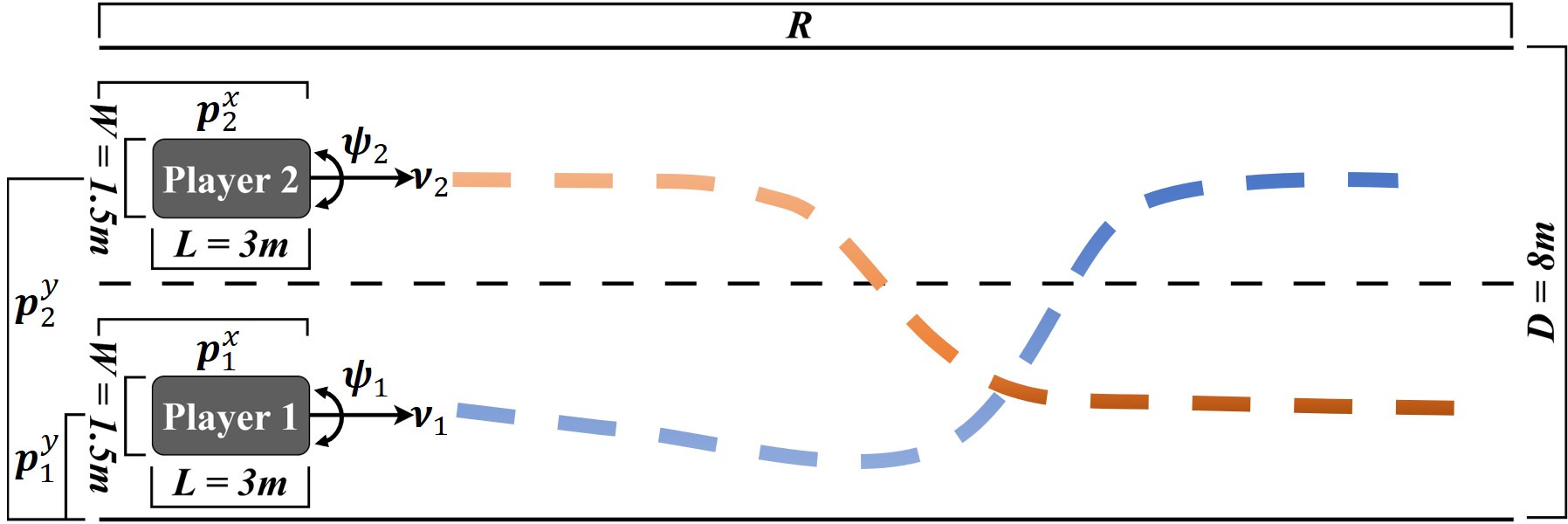}
\caption{Double-lane change setup with two players.}
\label{fig:lane change case}
\vspace{-10pt}
\end{figure}

\textbf{Experiment setup.} 
The schematic is shown in Figure~\ref{fig:lane change case}, depicting the states of Player $i$ as its location ($p^{x}_i$, $p^{y}_i$), orientation ($\psi_i$), and speed ($v_i$). $x_i := [p^{x}_i, p^{y}_i, \psi_i, v_i]^T$. 
The dashed blue and orange color (with increasing transparency along the x-axis) in the figure represents desired trajectories for both players. We use the same unicycle model and instantaneous loss as in Sec.~\ref{sec:narrow-road}. The terminal loss is set to incentivize players to stay within their respective lanes and regain the nominal speed:
\begin{equation}
\begin{aligned}
g_i(x_i) = -\mu p^x_{i}(T) + (p^y_{i}(T)-\bar{p}^{y}_i)^2 +  \\
(v_{i}(T)-\bar{v})^2 +\kappa(\psi_{i}(T)-\bar{\psi})^2,
\end{aligned}
\label{eq:case 3_terminal loss}
\end{equation}
where $\mu = 10^{-6}$, $\kappa = 100$, $\bar{p}^{y}_1 = 6 m$ for player 1 and $\bar{p}^{y}_2 = 2 m$ for player 2, $\bar{v} = 18 m/s$, $\bar{\psi} = 0 rad$, and $T = 4s$.

\begin{figure}[!ht]
\vspace{-12pt}
\centering
\includegraphics[width=0.96\linewidth]{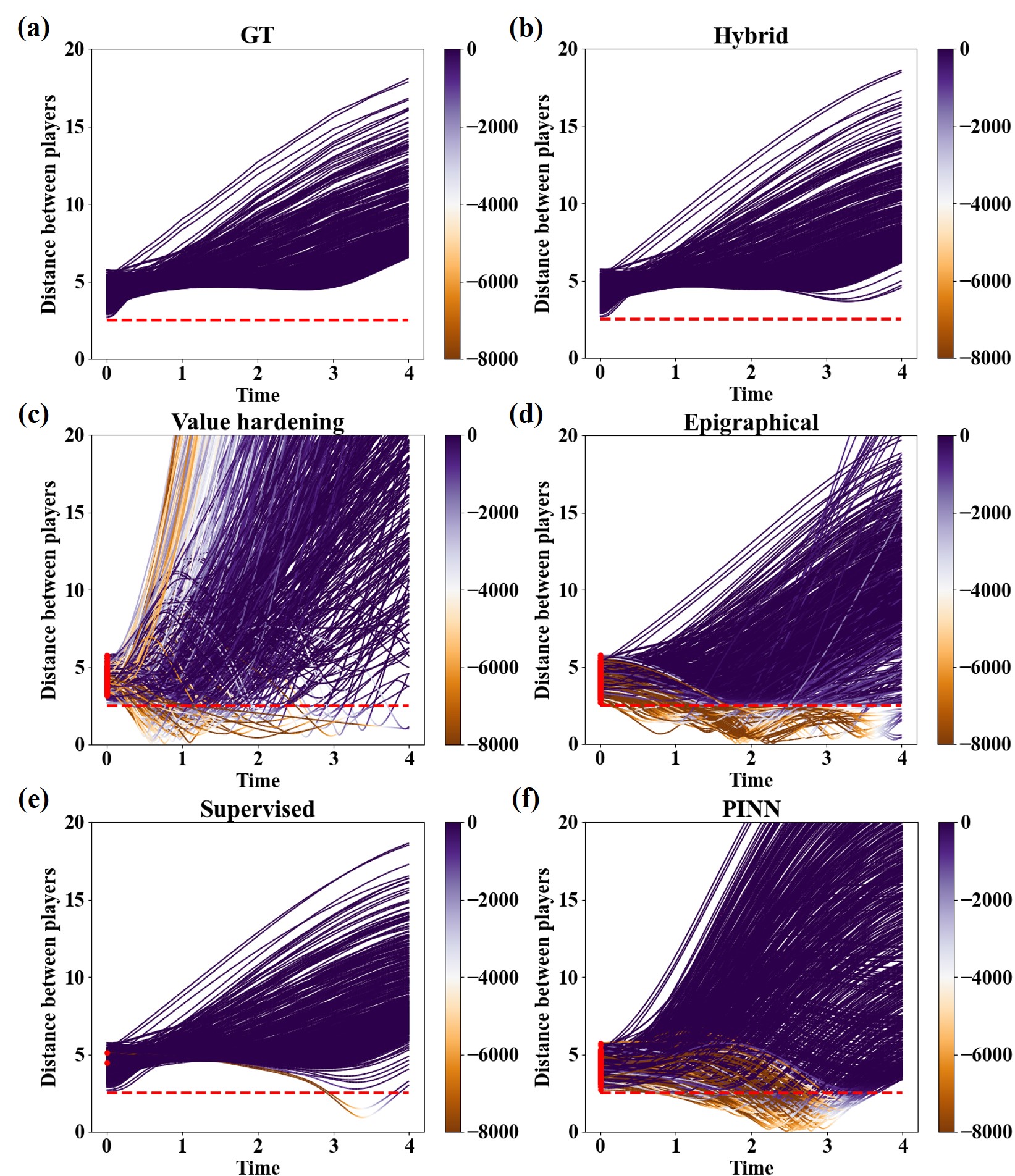}
\caption{(a): Ground truth distance between players over time for $\mathcal{X}_{GT}$. (b-e): Distance between players over time using hybrid, value hardening, epigraphical, supervised, and vanilla PINN models under $\mathcal{X}_{GT}$, respectively. Red dashed line represents the threshold distance for collision.}
\label{fig:lane_case}
\vspace{-10pt}
\end{figure}

\textbf{Data.} 
In the case of supervised learning, we generate 1.45k ground truth trajectories by uniformly sampling initial states from the set $\mathcal{X}_{GT}^1:= [0, 3]m \times [1.25, 2.75]m \times [-\pi/180, \pi/180] rad \times [18, 25]m/s$ for player 1, and $\mathcal{X}_{GT}^2:= [0, 3]m \times [5.25, 6.75]m \times [-\pi/180, \pi/180] rad \times [18, 25]m/s$ for player 2, resulting in a total of 118.9k data points. For vanilla and value-hardening PINN, we uniformly sample 162k states from the set $\mathcal{X}_{HJ}^1:= [0, 95]m \times [0, 6]m \times [-0.15, 0.13]rad \times [17, 26]m/s$ for player 1, and $\mathcal{X}_{HJ}^2:= [0, 95]m \times [2, 8]m \times [-0.13, 0.15]rad \times [17, 26]m/s$ for player 2. In the case of hybrid learning, we generate 1k ground truth trajectories (82k data points) by uniformly sampling initial states from $\mathcal{X}_{GT}^1$ for player 1, and $\mathcal{X}_{GT}^2$ for player 2. Additionally, we sample 80k states uniformly from $\mathcal{X}_{HJ}^1$ for player 1, and $\mathcal{X}_{HJ}^2$ for player 2. For epigraphical learning, we initially gather a sample of 200k states from $\mathcal{X}_{HJ}$ to ensure adherence to the boundary condition. We set the range of the auxiliary state $z_i$ as $[-9.5\times 10^{-5}, 400]$. All other settings follow Case 1.

\textbf{Training.} In this experiment, we employ the Adam optimizer with a constant learning rate of $1 \times 10^{-4}$. For vanilla PINN, we initiate the pre-training phase with 10k iterations, utilizing 162k boundary states uniformly sampled. Subsequently, we continue with the training phase, performing 350k iterations. For value hardening, we set the training duration for each hardening step to 7.2k iterations, completing a total of 50 steps to ensure a fair comparison. All other settings remain consistent with those of Case 1.

\begin{figure*}[!ht]
\centering
\includegraphics[width=1\linewidth]{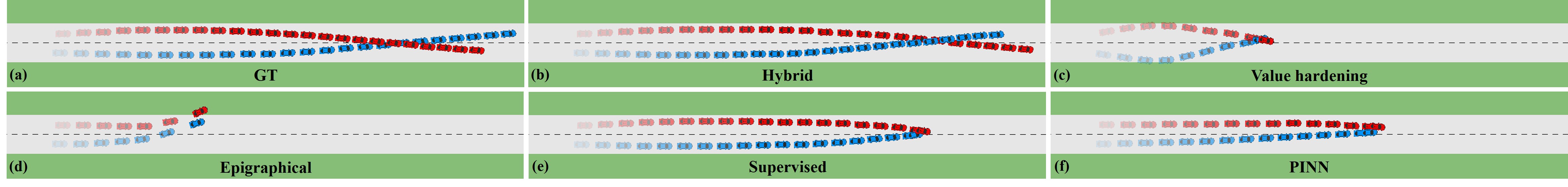}
\caption{Double-lane change visualization: (a): Ground truth safe trajectory. Transparency reduces along time. (b-e): Trajectories generated using hybrid, value hardening, epigraphical, supervised, and vanilla PINN models, respectively.}
\label{fig:lane change_visualization}
\vspace{-20pt}
\end{figure*}

\textbf{Results.} 
We assess the safety performance on a test set comprising 600 ground truth collision-free trajectories. These trajectories are generated by sampling initial states from $\mathcal{X}_{GT}$. The results are summarized in Tab.~\ref{table:lane_safety}, while the interaction distances between players are visualized in Fig.~\ref{fig:lane_case}. Similar to Cases 1 and 2, the hybrid method demonstrates superior performance compared to the others. Similar to Case 2, value hardening fails to generalize effectively within a computational budget similar to hybrid learning. Fig.~\ref{fig:lane change_visualization} shows interaction trajectories starting from one particular initial state where the hybrid method achieves safe interaction while the others fail.

\begin{table}[!ht]
\vspace{-5pt}
\centering
    \caption{Collision rate w/ different activation functions}
    \label{table:lane_safety}
\begin{tabular}{lcccccc}
\toprule
\multirow{2}{*}{Test} &
\multirow{2}{*}{Activation} & & \multicolumn{3}{c}{Learning Method}  \\ \cmidrule(lr){3-7}  
Domain & Functions
& \textbf{HL} & \textbf{VH} & \textbf{EL} & \textbf{SL}  & \textbf{PINN}\\ \midrule
& \texttt{tanh}  & \textbf{0.00}\%  & 23.0\% & 46.2\% & 0.33\% & 30.2\% \\
$\mathcal{X}_{GT}$ & \texttt{relu}  & 1.33\% & 40.3\%  & 61.0\% & \textbf{0.00}\%  & 52.5\% \\
& \texttt{sine}  & \textbf{0.50}\% & 11.2\% & 48.5\% & 1.00\%  & 17.3\% \\ 
\bottomrule
\end{tabular}
\vspace{-5pt}
\end{table}

\vspace{-0.2in}
\subsection{Case 4: two-drone collision avoidance}
\label{sec:drone-avoidance}

\begin{figure}[!ht]
\centering
\includegraphics[width=1\linewidth]{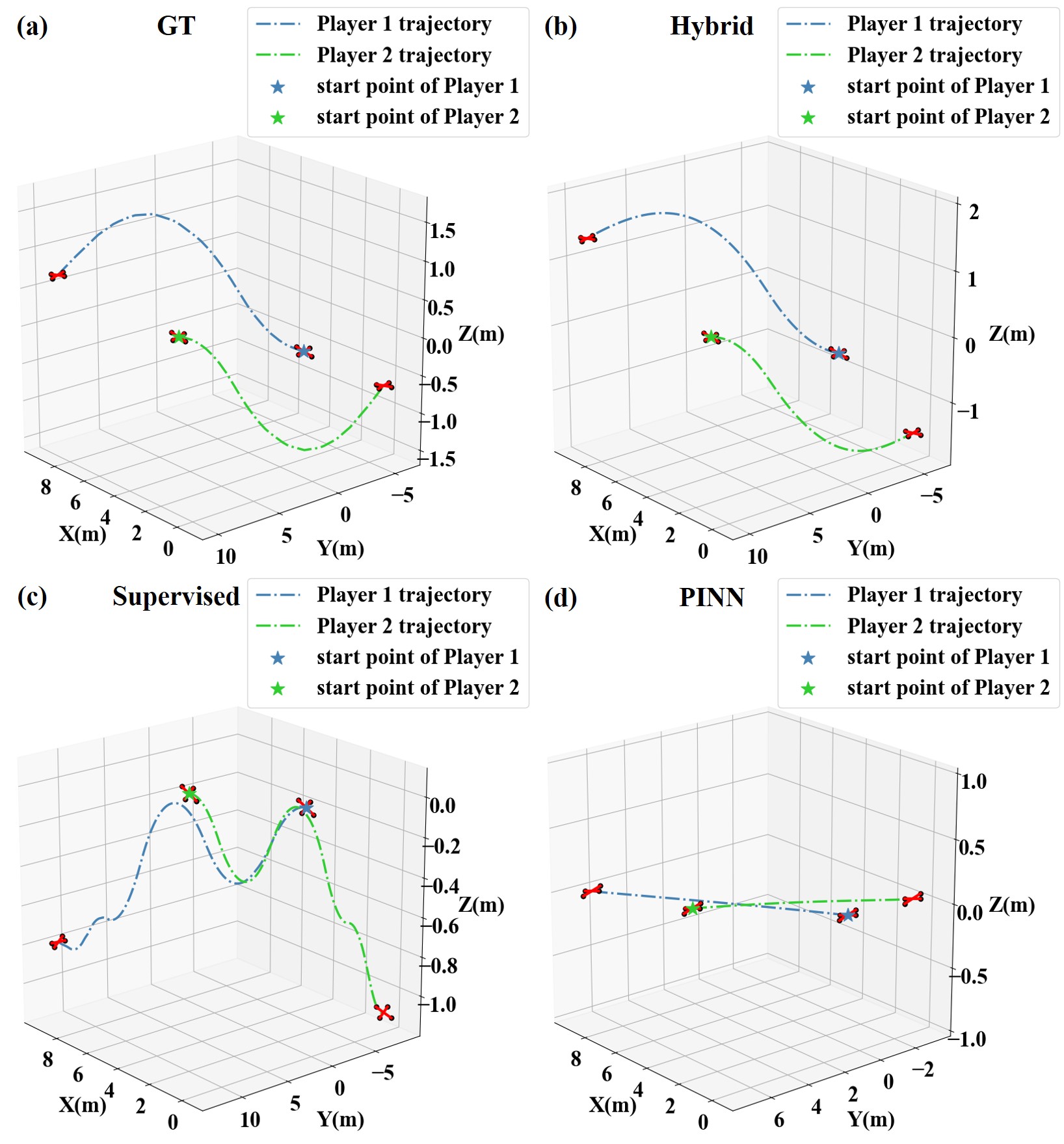}
\caption{Two-drone collision avoidance visualization: (a): Ground truth safe trajectory. (b-d): Trajectories generated using hybrid, supervised, and vanilla PINN models, respectively.}
\label{fig:drone_visualization}
\vspace{-20pt}
\end{figure}

\textbf{Experiment setup.}
In this experiment, we consider that the states of Player $i$ consist of its location ($p^{x}_i$, $p^{y}_i$, $p^{z}_i$), and speed ($v^{x}_i$, $v^{y}_i$, $v^{z}_i$), denoted as $x_i := [p^{x}_i, p^{y}_i, p^{z}_i, v^{x}_i, v^{y}_i, v^{z}_i]^T$. We use the flight dynamics (in the near-hover regime, at zero yaw with respect to a global coordinate frame) described in~\cite{fridovich2020confidence}:
\begin{eqnarray}
\begin{aligned}
\left[
\begin{array}{c}
    \dot p_i^x \\
    \dot p_i^y \\
    \dot p_i^z \\
    \dot v_i^x \\
    \dot v_i^y \\
    \dot v_i^z \\
\end{array}
\right]
=
\left[
\begin{array}{c}
    v_i^x \\
    v_i^y \\
    v_i^z \\
    g \tan(\theta_i) \\
    -g \tan(\phi_i) \\
    \tau_i - g \\
\end{array}
\right],
\end{aligned}
\end{eqnarray} 
where the tracking control $u_i = (\theta_i, \phi_i, \tau_i)$ corresponds to roll, pitch and thrust. In this experiment, $\theta_i \in [-0.05, 0.05] rad$, $\phi_i \in [-0.05, 0.05] rad$, $\tau_i \in [7.81, 11.81] m/s^2$, and $g = 9.81 m/s^2$. Note that we have assumed a zero yaw angle for the quadrotor. The instantaneous loss considers the control effort and the collision penalty:
\begin{equation}
\begin{aligned}
    {\tilde l_i}(\textbf{x}_i, \omega_i, u_i) = k_{\theta}\tan^2(\theta_i) + k_{\phi}\tan^2(\phi_i) \\
    + (\tau_i - g)^2 + b\sigma(\textbf{x}_i,\eta),
\end{aligned}
\end{equation}
where the penalty function is defined as 
\begin{equation}
\begin{aligned}
    & \sigma(\textbf{x}_i,\eta) = \left(1+ \exp(\gamma (S - \eta)\right))^{-1}, \notag \\
    & S = \sqrt{((R_x - p_2^x) - p_1^x)^2  + ((R_y - p_2^y) - p_1^y)^2 + (p_2^z - p_1^z)^2}. \notag
\end{aligned}
\end{equation}
$b = 10^4$ and $\gamma=5$.
Additionally, the parameters $R_x=5m$ and $R_y=5m$ are used to transform the coordinate positions of the two players along the $x-$ and $y-$ axes, respectively. The values of $k_{\theta}=100$ and $k_{\phi}=100$ determine the trade-off between control effort for roll, pitch, and thrust. Furthermore, $\eta=0.9 m$ represents the collision threshold. The terminal loss is set to encourage players to move along their respective $x$ and $y$ directions, to return to $0m$ on the $z$ axis, and to remain stationary when the simulation is complete:
\begin{equation}
\begin{aligned}
g_i(x_i) = -\mu p^x_i(T) - \mu p^y_i(T) + (p^z_{i}(T)-\bar{p}^{z}_i)^2 +  \\
(v^x_i(T) - \bar{v}^x_i)^2 + (v^y_i(T) - \bar{v}^y_i)^2 + (v^z_i(T) - \bar{v}^z_i)^2.
\end{aligned}
\end{equation}
where $\mu = 10^{-6}$, $\bar{p}^z_i = 0 m$ , $\bar{v}^x_i = \bar{v}^y_i = \bar{v}^z_i = 0 m/s$, and $T = 4s$. In this case study, we only compare the generalization and safety performance between the hybrid and the supervised methods, and use vanilla PINN as a baseline. Value hardening and epigraphical learning are dropped from the comparison since they do not generalize well in high-dimensional cases as we found in Sec.~\ref{sec:narrow-road} and \ref{sec:lane-change}.

\begin{figure}[!ht]
\vspace{-10pt}
\centering
\includegraphics[width=0.96\linewidth]{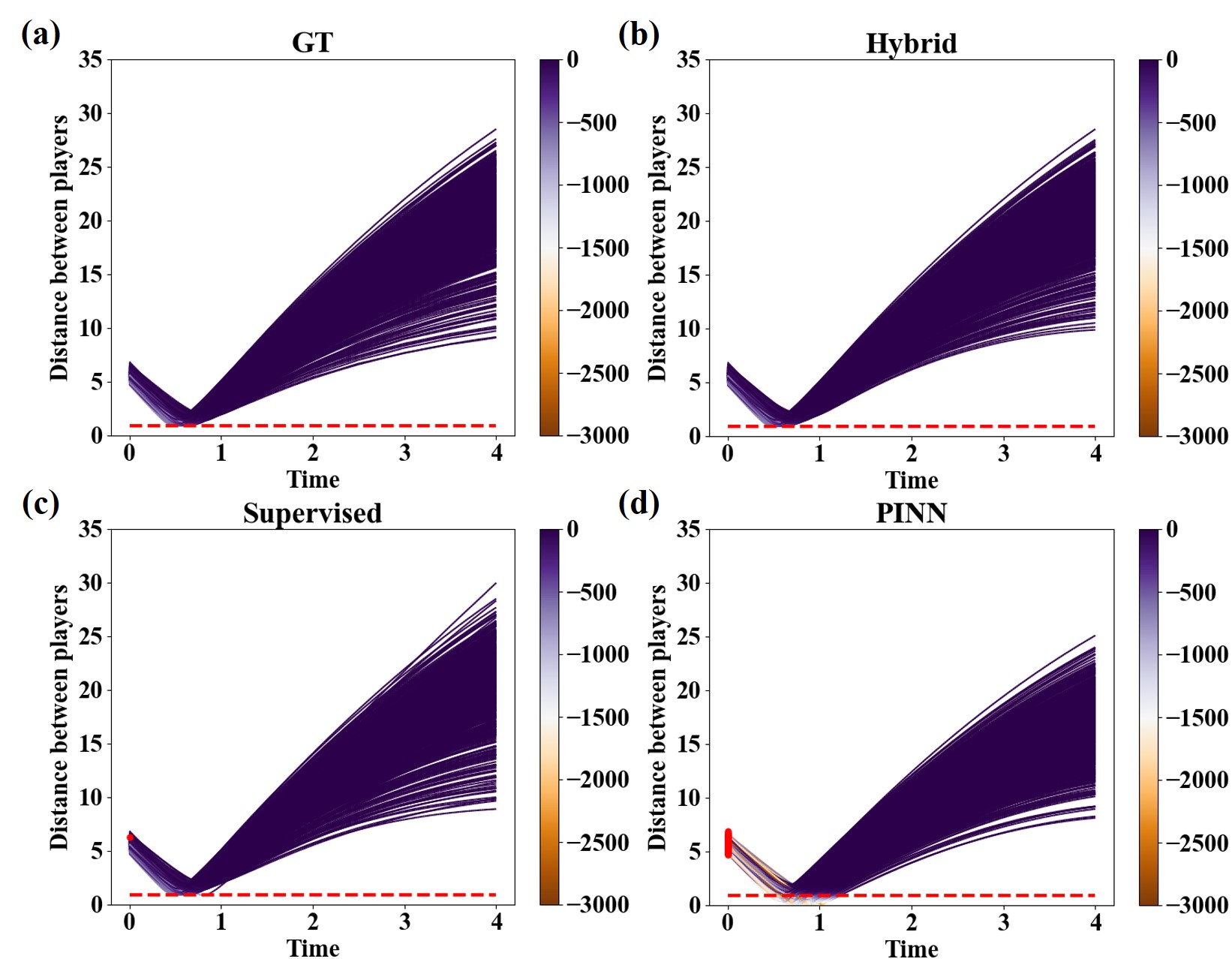}
\caption{(a): Ground truth distance between players over time for $\mathcal{X}_{GT}$. (b-d): Distance between players over time using hybrid, supervised, and vanilla PINN models under $\mathcal{X}_{GT}$, respectively. Red dashed line represents the threshold distance for collision.}
\label{fig:drone_case}
\vspace{-15pt}
\end{figure}

\textbf{Data.} 
In the case of supervised learning, we generate 1.25k ground truth trajectories by uniformly sampling initial states from the set $\mathcal{X}_{GT}:= [0, 1]m \times [0, 1]m \times [-0.1, 0.1]m \times [2, 4]m/s \times [2, 4]m/s \times [0, 0.1]m/s$, resulting in a total of 102.5k data points. For vanilla PINN, we uniformly sample 162k states from the set $\mathcal{X}_{HJ}:= [0, 15.5]m \times [0, 15.5]m \times [-1.8, 2]m \times [0.3, 4.5]m/s \times [0.3, 4.5]m/s \times [-1.8, 1.8]m/s$. In the case of hybrid learning, we generate 1k ground truth trajectories (82k data points) by uniformly sampling initial states from 
$\mathcal{X}_{GT}$. Additionally, we sample 80k states uniformly from $\mathcal{X}_{HJ}$.

\textbf{Training.} We use the Adam optimizer with a fixed learning rate of $1 \times 10^{-4}$. For vanilla PINN, we pretrain the networks for 100k iterations using 162k uniformly sampled boundary states and subsequently train them for an additional 400k iterations. The remaining settings for this experiment align with those used in Case 1.

\begin{table}[!ht]
\centering
    \caption{Collision rate w/ different activation functions}
    \label{table:drone_safety}
\begin{tabular}{lcccc}
\toprule
\multirow{2}{*}{Test} &
\multirow{2}{*}{Activation} & & \multicolumn{1}{c}{Learning Method}  \\ \cmidrule(lr){3-5}  
Domain & Functions
& \textbf{HL} & \textbf{SL}  & \textbf{PINN}\\ \midrule
& \texttt{tanh}  & \textbf{0.00}\% & 0.17\% & 75.8\% \\
$\mathcal{X}_{GT}$ & \texttt{relu}  & 34.0\% & \textbf{0.67}\%  & 76.2\% \\
& \texttt{sine}  & \textbf{0.00}\% & 0.17\% & 75.7\% \\ 
\bottomrule
\end{tabular}
\vspace{-5pt}
\end{table}

\textbf{Results.} 
We assess the safety performance on a test set comprising 600 ground truth collision-free trajectories. These trajectories are generated by uniformly sampling initial states from $\mathcal{X}_{GT}$. The results are summarized in Tab.~\ref{table:drone_safety}, while the interaction distances between players are visualized in Fig.~\ref{fig:drone_case}. Similar to the first three cases, the hybrid learning method demonstrates superior performance compared to the other methods. Fig.~\ref{fig:drone_visualization} visualizes the trajectories starting from a particular initial state where the hybrid method achieves safe interaction, while the other baselines yield collisions and undesired trajectories.

\vspace{-0.1in}
\section{Discussion}
\label{sec:discussion}
\subsection{Safety guarantee}
\label{sec:safety guarantee}
We note that our method does not provide safety certificate in its current form and discuss potential future directions. 
\textit{Policy certification:} For fixed-time differential games, it is possible to consider the interaction, i.e., the interchanging computation of actions (via approximated value gradients) and states (via an ODE solver), as a neural-network controlled system (NNCS), for which certification tools emerge~\cite{huang2019reachnn,manzanas2022reachability}. It should be noted that reachability analysis of NNCS is currently limited to small state space (due to the exponential growth in the approximation polynomial degree with respect to the state space dimensionality~\cite{huang2019reachnn}), small Lipschitz constant (due to linear growth of approximation error with respect to the Lipschitz constant of the neural network), and small network sizes (e.g., 4 layers each with 20 neurons in \cite{huang2019reachnn}). {Specifically, reachability analysis (e.g., forward~\cite{hu2020reach}, backward~\cite{everett2021reachability}, or automated~\cite{entesari2023automated} methods) for NNCS can be applied to a 6D quadrotor system. However, these analyses are limited to small network sizes and face challenges in achieving real-time verification for each closed-loop policy using trained models.}
\textit{Post-hoc state-constrained control:} When policy certification becomes intractable, an alternative could be to use a linear-quadratic reformulation of the game with conservative state constraint approximation for online computation of policies. In this setting, the value approximation network offers good initial policy guesses. This method trades off overall performance of policies in attaining Nash equilibrium for a computationally tractable safety guarantee. {A recent study~\cite{lin2023generating} explores online value approximations with safety guarantees for zero-sum games, yet it does not cover general-sum games and safe reachable analysis of online policy computation. \cite{hsu2023safety} proposes a unified framework to review the existing safety analysis approaches for closed-loop policy.}

\vspace{-0.15in}
\subsection{Consistency between BVP and HJI values} 
\label{sec:numerical value comparison}
Recognizing that PMP is only necessary conditions for local optimality~\cite{bressan2010noncooperative} while HJ solutions satisfy global optimality, we adopted multiple initial guesses to solve BVPs in order to seek global solutions. This treatment was applied to all case studies. Taking Case 1 as an example, we initialize the BVP solver with four state trajectories that follow constant control inputs: $\{(-5,-5)m/s^2, (-5,10)m/s^2, (10,-5)m/s^2, (10,10)m/s^2\}$. These trajectories represent four categories of interactions where each of the players either yield or accelerate through the intersection, and potentially lead to different equilibria. To address the issue with multiplicity of equilibrium, we choose the one that yields the best sum of values (i.e., Pareto optimal Nash equilibrium).

\begin{figure}[!ht]
\vspace{-10pt}
\centering
\includegraphics[width=0.96\linewidth]{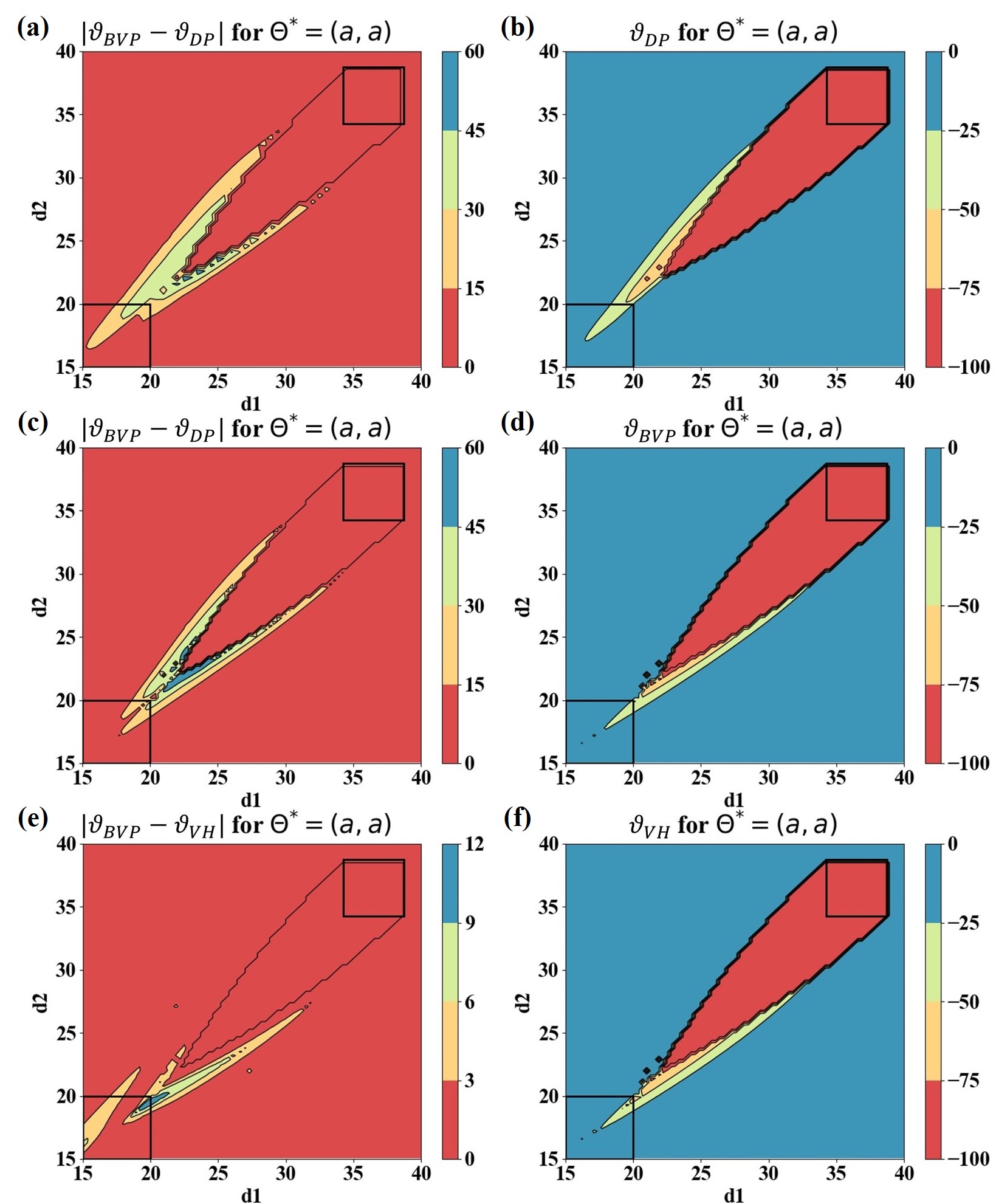}
\caption{(a,c): Difference $|\vartheta_{BVP}-\vartheta_{DP}|$ in $(d_1, d_2)$ frame with $v_{1,2}=18m/s$ at $t=0$ using DP spatial resolution $dx=0.5$ and $dx=0.3$, respectively. (b,d): Numerical solutions $\vartheta$ obtained through DP and BVP solver, respectively. (e): Difference $|\vartheta_{BVP}-\vartheta_{VH}|$ in $(d_1, d_2)$ frame with $v_{1,2}=18m/s$ at $t=0$. (f): Approximated solutions $\vartheta$ obtained through HJI-based learning approach-value hardening.}
\label{fig:BVP_HJI}
\vspace{-15pt}
\end{figure}

In the following, we empirically show that this treatment leads to consistent value landscapes between BVP and HJI. A visualization of the 
comparison uses the value contour from Case 1, projected to $(d_1, d_2)$ with fixed $v_{1,2}=18m/s$ and $t=0$ and with player types (\texttt{a}, \texttt{a}). See Fig.~\ref{fig:BVP_HJI}.

To compute values from the BVP solver, we sample initial states from $[15,40]m \times [15,40]m$ with fixed $v_{1,2}=18m/s$ and with the spatial resolution $dx=0.3m$. For each initial state, we solve the BVP and compute $\vartheta$ at $t=0$.

For HJI values, we consider two approximations of the ground truth. First, we extend an existing HJ PDE solver~\cite{bui2022optimizeddp} from zero-sum to general-sum. Since this DP solver has limited scalability with respect to state dimensions (up to 6D as demonstrated in \cite{bui2022optimizeddp}), we only applied the solver to Case 1 where values are 5D. The value difference between BVP and the DP solver, $|\vartheta_{BVP}-\vartheta_{DP}|$, is visualized for two DP spatial resolution settings: $dx=0.5$ and $dx=0.3$ in Fig.~\ref{fig:BVP_HJI}a and c, respectively. $dx=0.3$ is the highest resolution supported by our computing hardware. Note that values reported do not take into account constraint violation penalty, and unsafe states are assigned a constant value of $-100$. We observe that the difference $|\vartheta_{BVP}-\vartheta_{DP}|$ decreases as the resolution improves, and expect the trend to continue if the resolution were to be further increased. Since the DP solver has limited resolution, we resort to VH as a second approximation of the ground truth because it achieves relatively good generalization performance without using supervisory data in Case 1. Fig.~\ref{fig:BVP_HJI}e visualizes the value difference between BVP and VH, $|\vartheta_{BVP}-\vartheta_{VH}|$, which again shows the similarity between the two.

\vspace{-0.1in}
\subsection{Importance of the costate loss for safety performance}
\label{sec:EL explain}
\begin{figure}[!ht]
\vspace{-5pt}
\centering
\includegraphics[width=0.96\linewidth]{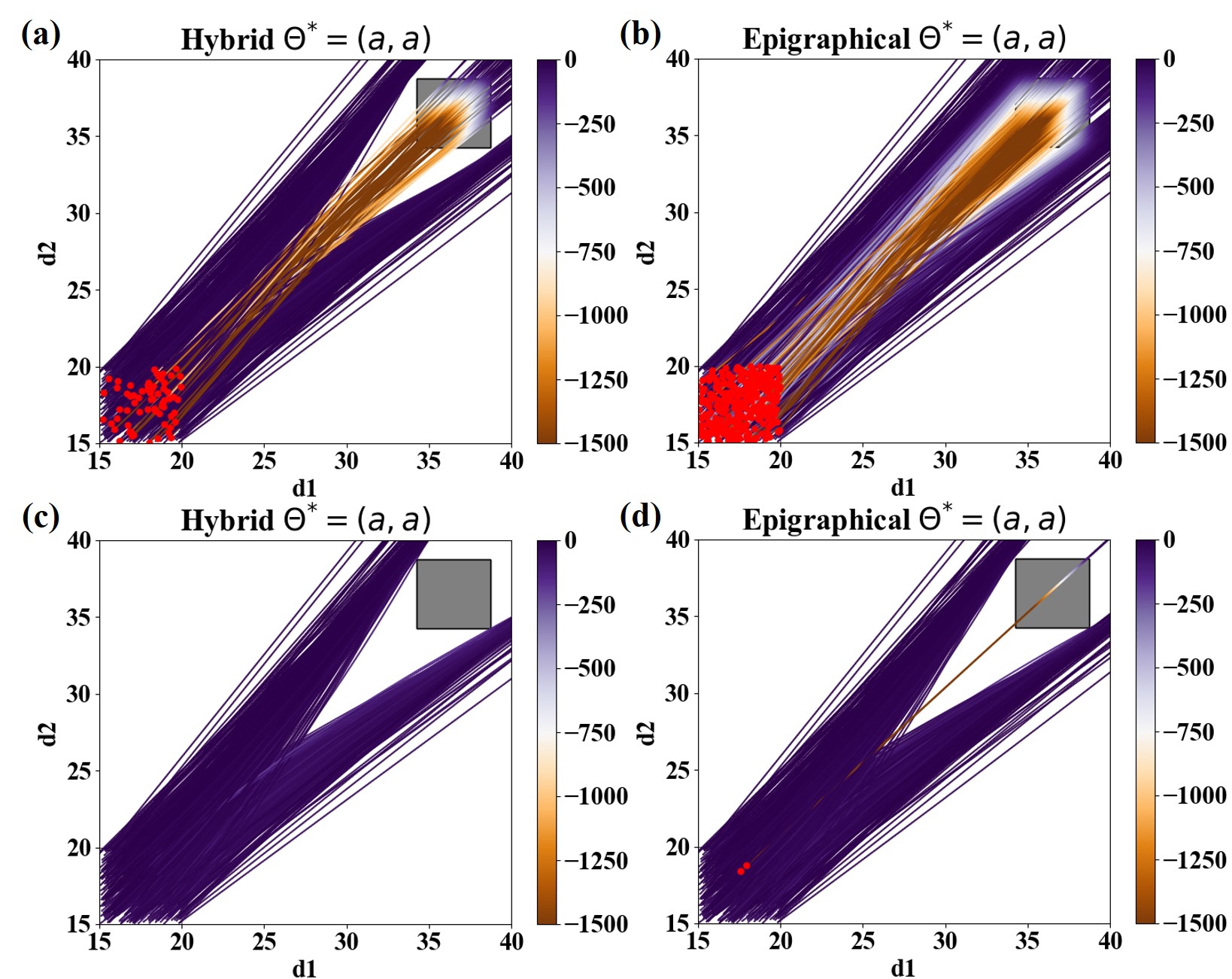}
\caption{(a,c): Trajectories generated using HL w/o and w costate loss under $\mathcal{X}_{GT}$. (b,d): Trajectories generated using EL w/o and w costate loss under $\mathcal{X}_{GT}$.}
\label{fig:el_explain}
\vspace{-15pt}
\end{figure}

Lastly, we provide details on the empirical study where we show that achieving good safety performance requires accuracy costate (value gradient) approximation. 
While the epigraphical technique facilitates smooth value approximation, it does not explicitly enforce small approximation errors on costates. In the following, we conducted a comparison between the hybrid learning (HL) and the epigraphical learning (EL) methods with identical training settings for Case 1: During their training, HL and EL uniformly sample 1k ground-truth trajectories (62k data points) in $\mathcal{X}_{GT}$ and 60k states in $\mathcal{X}_{HJ}$. Additionally, we uniformly sample the auxiliary state $z_i \in[-1.05\times 10^{-4}, 300]$ for EL. Both methods are solved using the Adam optimizer with a fixed learning rate of $2 \times 10^{-5}$. We pretrain the networks for 100k iterations using the supervised data and combine the supervised data with states sampled from an expanding time window starting from the terminal time to minimize $L_1 + L_2$ (Eq.~\eqref{eq:deepreach} and Eq.~\eqref{eq:supervised}) and $L_2+L_3$ (Eq.~\eqref{eq:supervised} and Eq.~\eqref{eq:hjpde_loss}) with 100k iterations for HL and EL, respectively. We show that the safety performance of EL is still worse than HL when using supervised data \textit{without} the costate loss in $L_2$ (see Fig.~\ref{fig:el_explain}b), and its safety performance significantly improves when the costate loss is considered (Fig.~\ref{fig:el_explain}d). On the other hand, Fig.~\ref{fig:el_explain}a shows that HL performs worse \textit{without} the costate loss. Hence, we conjecture that ensuring good safety performance requires not only small approximation errors for values but also for costates.

\vspace{-0.12in}
\section{Conclusion}
\label{sec:conclusion}
We proposed a hybrid learning method that combines the strengths of supervised learning and vanilla PINN to approximate discontinuous value functions as solutions to two-player general-sum differential games. The proposed method yields better generalization and safety performance than an array of baselines, including supervised learning, vanilla PINN, value hardening, and epigraphical learning, when using the same computational budget. We empirically demonstrate that the costate loss is the key factor for high safety performance, and the choice of the activation function and its parameters is crucial to the safety performance of learned models. Finally, all results in this paper can be reproduced using our code \href{https://github.com/dayanbatuofu/Value_Appro_Game}{https://github.com/dayanbatuofu/Value$\_$Appro$\_$Game}

\vspace{-0.1in}
\section*{Acknowledgments}
This work was in part supported by NSF CMMI-1925403 and NSF CNS-2101052. The views and conclusions contained in this document are those of the authors and should not be
interpreted as representing the official policies, either expressed or implied, of the National Science Foundation or the U.S. Government. 

\vspace{-0.2in}
\section{Appendix}
\label{sec:appendix}
\subsection{Summary of acronyms}
\label{sec:acronym}
\noindent
Tab.~\ref{table:acronym} summarizes acronyms used in the paper.
\begin{table}[!ht]
\vspace{-5pt}
\centering
    \caption{Acronyms used throughout the paper}
    \label{table:acronym}
\begin{tabularx}{1.03\linewidth}{lllX}
\toprule
{Acronym} & {Full Name} & {Acronym} & {Full Name}\\ 
\midrule
HJI & Hamilton-Jacobi-Isaacs & PINN & Physics-Informed Neural Network \\
CoD & Curse of Dimensionality & PMP & Pontryagin's Maximum Principle \\
DP & Dynamic Programming & NNCS & Neural-Network Control System\\
HRI & Human-Robot Interaction & \texttt{a} & aggressive\\
MPC & Model Predictive Control & \texttt{na} & non-aggressive\\
BVP & Boundary Value Problem & e & empathetic\\
ENO & Essentially Non-Oscillatory & ne & non-empathetic\\
MAEs & Mean Absolute Errors & GT & Ground Truth \\
EL & Epigraphical Learning & HL & Hybrid Learning\\
SL & Supervised Learning & VH & Value Hardening\\
\bottomrule
\end{tabularx}
\vspace{-5pt}
\end{table}

\vspace{-0.2in}
\subsection{Proof of Lemma 1 (Following proofs in~\cite{lee2022safety})}
\label{sec:Lemma 1}
\begin{proof}
\noindent
(i) $\vartheta_i(\textbf{x}_i, t)-z_i \leq 0$ implies that there exists $\alpha_i \in \mathcal{A}$ such that
\begin{equation}
\begin{aligned}
\int_t^T l_i(x_s^{\textbf{x}_i,t, \alpha_i}, \alpha_i\left(\textbf{x}_s^{\textbf{x}_i,t,\alpha_i,\alpha_{-i}}, s\right))ds + g_i\left(x_T^{\textbf{x}_i,t, \alpha_i}\right)-z_i \leq 0,
\end{aligned}
\end{equation}
and $c_i(\textbf{x}_s^{\textbf{x}_i,t,\alpha_i,\alpha_{-i}}) \leq 0$ for $s \in [t, T]$. Thus, there exists $\alpha_i$ such that $V_i(\textbf{x}_i,z_i, t) \leq 0$.

(ii) $V_i(\textbf{x}_i,z_i,t) \leq 0$ and $c_i(\textbf{x}_s^{\textbf{x}_i,t,\alpha_i,\alpha_{-i}}) \leq 0$ implies that there exists $\alpha_i \in \mathcal{A}$ such that 
\begin{equation}
\begin{aligned}
\int_t^T l_i(x_s^{\textbf{x}_i,t,\alpha_i}, \alpha_i\left(\textbf{x}_s^{\textbf{x}_i,t,\alpha_i,\alpha_{-i}}, s\right))ds
+ g_i\left(x_T^{\textbf{x}_i,t,\alpha_i}\right)-z_i \leq 0.
\end{aligned}
\end{equation}
which concludes $\vartheta_i(\textbf{x}_i, t)-z_i \leq 0$.
\end{proof}

\vspace{-0.2in}
\subsection{Proof of Lemma 2 (Following proofs in~\cite{lee2022safety,evans2022partial})}
\label{sec:Lemma 2}
\begin{proof}
For any policy $\alpha_i$ and a small step $h>0$, we can use Eq.~\eqref{eq:big v_fn_g} to derive the following relation ($\alpha_{-i}^*$ represents equiliribum policy for the fellow player of Player $i$). 
\begin{equation}
\begin{aligned}
V_i(\textbf{x}_i,z_i,t) = &\min_{\alpha_{i} \in \mathcal{A}}\max\bigg\{\max_{s\in [t,T]}c_i(\textbf{x}_s^{\textbf{x}_i,t,\alpha_i,\alpha_{-i}^*}), \\ 
& g_i\left(x_T^{\textbf{x}_i,t,\alpha_i}\right) - z_i(T)\bigg\}, \notag \\
= &\max\bigg\{\max_{s\in [t,t+h]}c_i(\textbf{x}_s^{\textbf{x}_i,t,\alpha_i,\alpha_{-i}^*}), \\
&\max\{\max_{s\in [t+h,T]}c_i(\textbf{x}_s^{\textbf{x}_i,t,\alpha_i,\alpha_{-i}^*}), \\
& g_i\left(x_T^{\textbf{x}_i,t,\alpha_i}\right) - z_i(T)\}\bigg\}.
\end{aligned}
\end{equation}
There exists two different policies $\alpha_{i_1}, \alpha_{i_2} \in \mathcal{A}$ such that
$$ \alpha_i=\left\{
\begin{aligned}
&\alpha_{i_1}(s), \quad s \in [t, t+h], \\
&\alpha_{i_2}(s), \quad s \in (t+h, T]. \\
\end{aligned}
\right.
$$
Then we have
\begin{equation}
\begin{aligned}
V_i(\textbf{x}_i,z_i,t) = &\min_{\alpha_{i_1} \in \mathcal{A}, \alpha_{i_2} \in \mathcal{A}}\max\bigg\{\max_{s\in [t,t+h]}c_i(\textbf{x}_s^{\textbf{x}_i,t,\alpha_i,\alpha_{-i}^*}), \\
& \max\{\max_{s\in [t+h,T]}c_i(\textbf{x}_s^{\textbf{x}_i,t,\alpha_i,\alpha_{-i}^*}), \\
& g_i\left(x_T^{\textbf{x}_i,t,\alpha_i}\right) - z_i(T)\}\bigg\}, \\
= &\min_{\alpha_{i_1} \in \mathcal{A}}\max\bigg\{\max_{s\in [t,t+h]}c_i(\textbf{x}_s^{\textbf{x}_i,t,\alpha_i,\alpha_{-i}^*}), \\
& \min_{\alpha_{i_2} \in \mathcal{A}}\max\{\max_{s\in [t+h,T]}c_i(\textbf{x}_s^{\textbf{x}_i,t,\alpha_i,\alpha_{-i}^*}), \\
& g_i\left(x_T^{\textbf{x}_i,t,\alpha_i}\right) - z_i(T)\}\bigg\}, \\
= &\min_{\alpha_{i_1} \in \mathcal{A}}\max\bigg\{\max_{s\in [t,t+h]}c_i(\textbf{x}_s^{\textbf{x}_i,t,\alpha_i,\alpha_{-i}^*}), \\
& V_i(\textbf{x}_i(t+h), z_i(t+h), t+h)\bigg\}, \\
= &\min_{\alpha_i \in \mathcal{A}}\max\bigg\{\max_{s\in [t,t+h]}c_i(\textbf{x}_s^{\textbf{x}_i,t,\alpha_i,\alpha_{-i}^*}), \\
& V_i(\textbf{x}_i(t+h), z_i(t+h), t+h)\bigg\}. \notag
\end{aligned}
\end{equation}
\end{proof}

\vspace{-0.4in}
\subsection{Proof of Theorem 1 (Following proofs in~\cite{lee2022safety,evans2022partial})}
\label{sec:Theorem 1}
\begin{proof}
(i) When $t=T$, $V_i$ is easily satisfied based on definition 
\begin{equation}
\begin{aligned}
V_i(\textbf{x}_i,z_i, T) & = \max\left\{c_i(\textbf{x}_T^{\textbf{x}_i,t,\alpha_i,\alpha_{-i}^*}), g_i\left(x_T^{\textbf{x}_i,t,\alpha_i}\right) - z_i(T)\right\} \\
& = \max\left\{c_i(\textbf{x}_i(T)), g_i(T) - z_i(T)\right\}
\end{aligned}
\end{equation}

(ii) Let $W_i \in C^{\infty}(\mathcal{X} \times \mathbb{R} \times [0,T])$, and assume that $V_i-W_i$ has local maximum at $(\textbf{x}_i(t_0), z_i(t_0), t_0) \in \mathcal{X} \times \mathbb{R} \times [0, T)$ and $(V_i-W_i)(\textbf{x}_i(t_0), z_i(t_0), t_0) = 0$, we need to prove
\begin{equation}
\begin{aligned}
\max\big\{&c_i(\textbf{x}_{t_0}^{\textbf{x}_i,t,\alpha_i,\alpha_{-i}^*})-W_i(\textbf{x}_i(t_0), z_i(t_0), t_0),  \\
& \nabla_{t} W_i(\textbf{x}_i(t_0), z_i(t_0), t_0) - \mathcal{H}_i(t_0, \textbf{x}_i(t_0), z_i(t_0),  \\
& \nabla_{\textbf{x}_i}W_i(\textbf{x}_i(t_0), z_i(t_0), t_0), \nabla_{z_i} W_i(\textbf{x}_i(t_0), z_i(t_0), t_0))\big\} \\
& \geq 0.
\end{aligned}
\end{equation}
Suppose not. Then there exists $\xi > 0$ and $\tilde\alpha_{i} \in \mathcal{A}$ such that 
\begin{equation}
\begin{aligned}
& c_i(\textbf{x}_s^{\textbf{x}_i,t,\tilde\alpha_i, \alpha_{-i}^*})-W_i(\textbf{x}_i(t_0), z_i(t_0), t_0) \leq -\xi, \\
& \nabla_{t} W_i(\textbf{x}_i,z_i,t) + \nabla_{\textbf{x}_i}W_i(\textbf{x}_i,z_i,t)\cdot \textbf{f}_i(\textbf{x}_i,\tilde\alpha_{i}, \alpha_{-i}^*)  \\ 
& - \nabla_{z_i} W_i(\textbf{x}_i,z_i,t)\cdot l_i\left(x_s^{\textbf{x}_i,t,\tilde\alpha_{i}}, \tilde\alpha_i\left(\textbf{x}_s^{\textbf{x}_i,t,\tilde\alpha_{i},\alpha_{-i}^*}, s\right)\right) \leq -\xi.
\end{aligned}
\end{equation}
for all points $(\textbf{x}_i,z_i, t)$ sufficiently close to $(\textbf{x}_i(t_0), z_i(t_0), t_0)$: there exists small enough $h_1 > 0$ such that $||\textbf{x}_i-\textbf{x}_i(t_0)||+|z_i-z_i(t_0)|+|t-t_0|<h_1$. According to Assumptions in Sec.~\ref{sec:assumptions}, choose a small $h$ such that $||\textbf{x}_i-\textbf{x}_i(t_0)||+|z_i-z_i(t_0)|<h_1-h$ for $s \in [t_0, t_0+h]$, then
\begin{equation}
\begin{aligned}
& c_i(\textbf{x}_s^{\textbf{x}_i,t,\tilde\alpha_i, \alpha_{-i}^*})-W_i(\textbf{x}_i(t_0), z_i(t_0), t_0) \leq -\xi, \\
& \nabla_{t} W_i(\textbf{x}_i,z_i,s) + \nabla_{\textbf{x}_i}W_i(\textbf{x}_i,z_i,s)\cdot \textbf{f}_i(\textbf{x}_i,\tilde\alpha_{i}, \alpha_{-i}^*)  \\
& - \nabla_{z_i} W_i(\textbf{x}_i,z_i,s)\cdot l_i\left(x_s^{\textbf{x}_i,t,\tilde\alpha_{i}}, \tilde\alpha_i\left(\textbf{x}_s^{\textbf{x}_i,t,\tilde\alpha_{i},\alpha_{-i}^*}, s\right)\right) \leq -\xi.
\label{eq:c_inequlaity_g}
\end{aligned}
\end{equation}

\noindent
According to the condition that $V_i-W_i$ has a local maximum at $(\textbf{x}(t_0), z_i(t_0), t_0)$, then
\begin{equation}
\begin{aligned}
& V_i(\textbf{x}_i(t_0+h), z_i(t_0+h), t_0+h)  \\
& - W_i(\textbf{x}_i(t_0+h), z_i(t_0+h), t_0+h)  \\
& \leq V_i(\textbf{x}_i(t_0), z_i(t_0), t_0) - W_i(\textbf{x}(t_0), z_i(t_0), t_0) \\
\Rightarrow & V_i(\textbf{x}_i(t_0+h), z_i(t_0+h), t_0+h) - V_i(\textbf{x}_i(t_0), z_i(t_0), t_0) \\
& \leq W_i(\textbf{x}_i(t_0+h), z_i(t_0+h), t_0+h) - W_i(\textbf{x}_i(t_0), z_i(t_0), t_0)  \\
\Rightarrow & V_i(\textbf{x}_i(t_0+h), z_i(t_0+h), t_0+h) - V_i(\textbf{x}_i(t_0), z_i(t_0), t_0)  \\
& \leq \int_{t_0}^{t_0+h}\frac{dW_i}{dt}ds  \\
\Rightarrow & V_i(\textbf{x}_i(t_0+h), z_i(t_0+h), t_0+h) - V_i(\textbf{x}_i(t_0), z_i(t_0), t_0)  \\
& \leq \int_{t_0}^{t_0+h}\bigg\{\nabla_{t} W_i(\textbf{x}_i,z_i,s) + \nabla_{\textbf{x}_i}W_i(\textbf{x}_i,z_i,s)\cdot \textbf{f}_i  \\
&  - \nabla_{z_i} W_i(\textbf{x}_i,z_i,s) \cdot l_i\bigg\}ds \leq -\xi h 
\label{eq:V_inequlaity_1_g}
\end{aligned}
\end{equation}
Lemma~\ref{lemma:lemma2} says that 
\begin{equation}
\begin{aligned}
V_i(\textbf{x}_i(t_0), z_i(t_0), t_0) = \min_{u_i \in \mathcal{U}_i}\max\bigg\{\max_{s\in [t_0,t_0+h]}c_i(\textbf{x}_i(s)), \\
V_i(\textbf{x}_i(t_0+h), z_i(t_0+h), t_0+h)\bigg\},
\label{eq:optimality_g}
\end{aligned}
\end{equation}
Subtract Eq.~\eqref{eq:optimality_g} by $W_i(\textbf{x}_i(t_0), z_i(t_0), t_0)$ on both sides and combine Eq.~\eqref{eq:c_inequlaity_g} and~\eqref{eq:V_inequlaity_1_g}:
\begin{equation}
\begin{aligned}
0 & = (V_i-W_i)(\textbf{x}_i(t_0), z_i(t_0), t_0) \\  
& = \min_{u_i \in \mathcal{U}_i}\max\left\{-\xi, -\xi h\right\} < 0,
\end{aligned}
\end{equation}
which is a contradiction. Thus, we prove that 
\begin{equation}
\begin{aligned}
\max\big\{&c_i(\textbf{x}_{t_0}^{\textbf{x}_i,t,\alpha_i,\alpha_{-i}^*})-W_i(\textbf{x}_i(t_0), z_i(t_0), t_0),  \\
& \nabla_{t} W_i(\textbf{x}_i(t_0), z_i(t_0), t_0) - \mathcal{H}_i(t_0, \textbf{x}_i(t_0), z_i(t_0),  \\
& \nabla_{\textbf{x}_i}{W_i(\textbf{x}_i(t_0), z_i(t_0), t_0)}, \nabla_{z_i} W_i(\textbf{x}_i(t_0), z_i(t_0), t_0))\big\} \\
& \geq 0.
\end{aligned}
\end{equation}

\noindent
(iii) Let $W_i \in C^{\infty}(\mathcal{X} \times \mathbb{R} \times [0,T])$, and assume that $V_i-W_i$ has local minimum at $(\textbf{x}_i(t_0), z_i(t_0), t_0)) \in \mathcal{X} \times \mathbb{R} \times [0, T)$ and $(V_i-W_i)(\textbf{x}_i(t_0), z_i(t_0), t_0)) = 0$, we need to prove
\begin{equation}
\begin{aligned}
\max\big\{&c_i(\textbf{x}_{t_0}^{\textbf{x}_i,t,\alpha_i,\alpha_{-i}^*})-W_i(\textbf{x}_i(t_0), z_i(t_0), t_0),  \\
& \nabla_{t} W_i(\textbf{x}_i(t_0), z_i(t_0), t_0) - \mathcal{H}_i(t_0, \textbf{x}_i(t_0), z_i(t_0),  \\
& \nabla_{\textbf{x}_i}{W_i(\textbf{x}_i(t_0), z_i(t_0), t_0)}, \nabla_{z_i} W_i(\textbf{x}_i(t_0), z_i(t_0), t_0))\big\} \\
& \leq 0.
\end{aligned}
\end{equation}
The definition of $V_i$ says that  
\begin{equation}
\begin{aligned}
V_i(\textbf{x}_i(t_0),z_i(t_0),t_0) = & \max\bigg\{\max_{s\in [t_0,T]}c_i(\textbf{x}_s^{\textbf{x}_i,t,\alpha_i,\alpha_{-i}^*}), \\
& g_i\left(x_T^{\textbf{x}_i,t,\alpha_i}\right) - z_i(T)\bigg\}, \\
\geq & \max\bigg\{c_i(\textbf{x}_{t_0}^{\textbf{x}_i,t,\alpha_i,\alpha_{-i}^*}), \\
& g_i\left(x_T^{\textbf{x}_i,t,\alpha_i}\right) - z_i(T)\bigg\}.
\label{eq:value_inequality}
\end{aligned}
\end{equation}
for all $\alpha_i \in \mathcal{A}(t_0)$. Subtract Eq.~\eqref{eq:value_inequality} by  $W_i(\textbf{x}_i(t_0), z_i(t_0), t_0)$ on both sides to have 
\begin{equation}
\begin{aligned}
0 = & (V_i-W_i)(\textbf{x}_i(t_0), z_i(t_0), t_0) \geq \max\{c_i(\textbf{x}_{t_0}^{\textbf{x}_i,t,\alpha_i,\alpha_{-i}^*}) - \\ 
& W_i(\textbf{x}_i(t_0), z_i(t_0), t_0), g_i\left(x_T^{\textbf{x}_i,t,\alpha_i}\right) - z_i(T) - \\
& W_i(\textbf{x}_i(t_0), z_i(t_0), t_0)\}. 
\end{aligned}
\end{equation}
Then we must prove the following inequality
\begin{equation}
\begin{aligned}
& \nabla_{t} W_i(\textbf{x}_i(t_0), z_i(t_0), t_0) - \mathcal{H}_i(t_0, \textbf{x}_i(t_0), z_i(t_0), \\ 
& \nabla_{\textbf{x}_i}W_i(\textbf{x}_i(t_0), z_i(t_0), t_0),\nabla_{z_i} W_i(\textbf{x}_i(t_0), z_i(t_0), t_0)) \leq 0,    
\end{aligned}
\end{equation}
Suppose not. Then there exists $\xi > 0$ such that 
\begin{equation}
\begin{aligned}
\nabla_{t} W_i(\textbf{x}_i,z_i,t) - \max_{u_i \in \mathcal{U}_i}\big[-\nabla_{\textbf{x}_i}W_i(\textbf{x}_i,z_i,t)\cdot \textbf{f}_i \\
+ \nabla_{z_i}W_i(\textbf{x}_i,z_i,t)\cdot l_i\big] \geq \xi,
\end{aligned}
\end{equation}
for all points $(\textbf{x}_i,z_i,t)$ sufficiently close to $(\textbf{x}_i(t_0), z_i(t_0), t_0)$: there exists small enough $h_1 > 0$ such that $||\textbf{x}_i-\textbf{x}_i(t_0)||+|z_i-z_i(t_0)|+|t-t_0|<h_1$. For any $\alpha_i \in \mathcal{A}$, where 
\begin{equation}
\begin{aligned}
\alpha_i \in & \argmax_{\alpha_i \in \mathcal{A}}-\nabla_{\textbf{x}_i}W_i(\textbf{x}_i,z_i,s)\cdot \textbf{f}_i(\textbf{x}_i,\alpha_{i}, \alpha_{-i}^*) \\
& + \nabla_{z_i} W_i(\textbf{x}_i,z_i,s)\cdot l_i\left(x_s^{\textbf{x}_i,t,\alpha_{i}}, \alpha_i\left(\textbf{x}_s^{\textbf{x}_i,t,\alpha_{i},\alpha_{-i}^*}, s\right)\right),
\end{aligned}
\end{equation}
According to Assumptions in Sec.~\ref{sec:assumptions}, choose a small $h$ such that $||\textbf{x}_i-\textbf{x}_i(t_0)||+|z_i-z_i(t_0)|<h_1-h$ for $s \in [t_0, t_0+h]$, then
\begin{equation}
\begin{aligned}
& \nabla_{t} W_i(\textbf{x}_i,z_i,s) + \nabla_{\textbf{x}_i}W_i(\textbf{x}_i,z_i,s)\cdot \textbf{f}_i(\textbf{x}_i,\alpha_{i}, \alpha_{-i}^*)  \\
& - \nabla_{z_i} W_i(\textbf{x}_i,z_i,s)\cdot l_i\left(x_s^{\textbf{x}_i,t,\alpha_{i}}, \alpha_i\left(\textbf{x}_s^{\textbf{x}_i,t,\alpha_{i},\alpha_{-i}^*}, s\right)\right) \geq \xi,
\label{eq:HJ_inequlaity_g}
\end{aligned}
\end{equation}
for all $s \in [t_0, t_0+h]$. We integrate Eq.~\eqref{eq:HJ_inequlaity_g} over $s \in [t_0, t_0+h]$ to get
\begin{equation}
\begin{aligned}
W_i(\textbf{x}_i(t_0+h), z_i(t_0+h), t_0+h) \\ 
- W_i(\textbf{x}_i(t_0), z_i(t_0), t_0) \geq \xi h,
\label{eq:U_inequlaity_g}
\end{aligned}
\end{equation}
We have the following relation because Eq.~\eqref{eq:U_inequlaity_g} holds for all $u_i \in \mathcal{U}_i$
\begin{equation}
\begin{aligned}
\min_{u_i \in \mathcal{U}_i} & W_i(\textbf{x}_i(t_0+h), z_i(t_0+h), t_0+h) \\
& - W_i(\textbf{x}_i(t_0), z_i(t_0), t_0) \geq \xi h,
\end{aligned}
\end{equation}
According to the condition that $V_i-W_i$ has a local minimum at $(\textbf{x}_i(t_0), z_i(t_0), t_0)$, then
\begin{equation}
\begin{aligned}
& \min_{u_i \in \mathcal{U}_i} V_i(\textbf{x}_i(t_0+h), z_i(t_0+h), t_0+h) \\
& - V_i(\textbf{x}_i(t_0), z_i(t_0), t_0) \\
& \geq \min_{u_i \in \mathcal{U}_i} W_i(\textbf{x}_i(t_0+h), z_i(t_0+h), t_0+h) \\ 
& - W_i(\textbf{x}_i(t_0), z_i(t_0), t_0) \\
& \geq  \xi h \\
\Rightarrow & \min_{u_i \in \mathcal{U}_i} V_i(\textbf{x}_i(t_0+h), z_i(t_0+h), t_0+h) \\ 
& > V_i(\textbf{x}_i(t_0), z_i(t_0), t_0)
\label{eq:V_inequlaity_2_g}
\end{aligned}
\end{equation}
However, Lemma~\ref{lemma:lemma2} says that 
\begin{equation}
\min_{u_i \in \mathcal{U}_i}V_i(\textbf{x}_i(t_0+h), z_i(t_0+h), t_0+h) \leq V_i(\textbf{x}(t_0), z_i(t_0), t_0),
\end{equation}
which is a contradiction. Thus, we prove that 
\begin{equation}
\begin{aligned}
\max\big\{&c_i(\textbf{x}_{t_0}^{\textbf{x}_i,t,\alpha_i,\alpha_{-i}^*})-W_i(\textbf{x}_i(t_0), z_i(t_0), t_0),  \\
& \nabla_{t} W_i(\textbf{x}_i(t_0), z_i(t_0), t_0) - \mathcal{H}_i(t_0, \textbf{x}_i(t_0), z_i(t_0),  \\
& \nabla_{\textbf{x}_i}{W_i(\textbf{x}_i(t_0), z_i(t_0), t_0)}, \nabla_{z_i} W_i(\textbf{x}_i(t_0), z_i(t_0), t_0))\big\} \\
& \leq 0.
\end{aligned}
\end{equation}
Hence, we prove that $V_i(\textbf{x}_i,z_i,t)$ is the viscosity solution. 
\end{proof}

\vspace{-0.2in}
\bibliography{journal}

\begin{thebibliography}{10}
\providecommand{\url}[1]{#1}
\csname url@samestyle\endcsname
\providecommand{\newblock}{\relax}
\providecommand{\bibinfo}[2]{#2}
\providecommand{\BIBentrySTDinterwordspacing}{\spaceskip=0pt\relax}
\providecommand{\BIBentryALTinterwordstretchfactor}{4}
\providecommand{\BIBentryALTinterwordspacing}{\spaceskip=\fontdimen2\font plus
\BIBentryALTinterwordstretchfactor\fontdimen3\font minus
  \fontdimen4\font\relax}
\providecommand{\BIBforeignlanguage}[2]{{%
\expandafter\ifx\csname l@#1\endcsname\relax
\typeout{** WARNING: IEEEtran.bst: No hyphenation pattern has been}%
\typeout{** loaded for the language `#1'. Using the pattern for}%
\typeout{** the default language instead.}%
\else
\language=\csname l@#1\endcsname
\fi
#2}}
\providecommand{\BIBdecl}{\relax}
\BIBdecl

\bibitem{duarte2018impact}
F.~Duarte and C.~Ratti, ``The impact of autonomous vehicles on cities: A
  review,'' \emph{Journal of Urban Technology}, vol.~25, no.~4, pp. 3--18,
  2018.

\bibitem{peters2018review}
B.~S. Peters, P.~R. Armijo, C.~Krause, S.~A. Choudhury, and D.~Oleynikov,
  ``Review of emerging surgical robotic technology,'' \emph{Surgical
  endoscopy}, vol.~32, pp. 1636--1655, 2018.

\bibitem{murphy2004human}
R.~R. Murphy, ``Human-robot interaction in rescue robotics,'' \emph{IEEE
  Transactions on Systems, Man, and Cybernetics, Part C (Applications and
  Reviews)}, vol.~34, no.~2, pp. 138--153, 2004.

\bibitem{leung2020infusing}
K.~Leung, E.~Schmerling, M.~Zhang, M.~Chen, J.~Talbot, J.~C. Gerdes, and
  M.~Pavone, ``On infusing reachability-based safety assurance within planning
  frameworks for human--robot vehicle interactions,'' \emph{The International
  Journal of Robotics Research}, vol.~39, no. 10-11, pp. 1326--1345, 2020.

\bibitem{gammoudi2023differential}
N.~Gammoudi and H.~Zidani, ``A differential game control problem with state
  constraints,'' \emph{Mathematical Control and Related Fields}, vol.~13,
  no.~2, pp. 554--582, 2023.

\bibitem{lewis2012optimal}
F.~L. Lewis, D.~Vrabie, and V.~L. Syrmos, \emph{Optimal control}.\hskip 1em
  plus 0.5em minus 0.4em\relax John Wiley \& Sons, 2012.

\bibitem{bellman1965dynamic}
R.~Bellman and R.~E. Kalaba, \emph{Dynamic programming and modern control
  theory}.\hskip 1em plus 0.5em minus 0.4em\relax Citeseer, 1965, vol.~81.

\bibitem{weinan2021algorithms}
E.~Weinan, J.~Han, and A.~Jentzen, ``{Algorithms for solving high dimensional
  PDEs: from nonlinear Monte Carlo to machine learning},'' \emph{Nonlinearity},
  vol.~35, no.~1, p. 278, 2021.

\bibitem{shin2020convergence}
Y.~Shin, J.~Darbon, and G.~E. Karniadakis, ``{On the convergence of physics
  informed neural networks for linear second-order elliptic and parabolic type
  PDEs},'' \emph{arXiv preprint arXiv:2004.01806}, 2020.

\bibitem{ito2021neural}
K.~Ito, C.~Reisinger, and Y.~Zhang, ``{A neural network-based policy iteration
  algorithm with global $H^{2}$-superlinear convergence for stochastic games on
  domains},'' \emph{Foundations of Computational Mathematics}, vol.~21, no.~2,
  pp. 331--374, 2021.

\bibitem{fuks2020limitations}
O.~Fuks and H.~A. Tchelepi, ``Limitations of physics informed machine learning
  for nonlinear two-phase transport in porous media,'' \emph{Journal of Machine
  Learning for Modeling and Computing}, vol.~1, no.~1, 2020.

\bibitem{mangasarian1966sufficient}
O.~L. Mangasarian, ``Sufficient conditions for the optimal control of nonlinear
  systems,'' \emph{SIAM Journal on control}, vol.~4, no.~1, pp. 139--152, 1966.

\bibitem{bengio2009curriculum}
Y.~Bengio, J.~Louradour, R.~Collobert, and J.~Weston, ``Curriculum learning,''
  in \emph{Proceedings of the 26th annual international conference on machine
  learning}, 2009, pp. 41--48.

\bibitem{altarovici2013general}
A.~Altarovici, O.~Bokanowski, and H.~Zidani, ``{A general Hamilton-Jacobi
  framework for non-linear state-constrained control problems},'' \emph{ESAIM:
  Control, Optimisation and Calculus of Variations}, vol.~19, no.~2, pp.
  337--357, 2013.

\bibitem{lee2022safety}
D.~Lee, ``Safety-guaranteed autonomy under uncertainty,'' Ph.D. dissertation,
  UC Berkeley, 2022.

\bibitem{raissi2019physics}
M.~Raissi, P.~Perdikaris, and G.~E. Karniadakis, ``Physics-informed neural
  networks: A deep learning framework for solving forward and inverse problems
  involving nonlinear partial differential equations,'' \emph{Journal of
  Computational physics}, vol. 378, pp. 686--707, 2019.

\bibitem{raissi2019deep}
M.~Raissi, Z.~Wang, M.~S. Triantafyllou, and G.~E. Karniadakis, ``Deep learning
  of vortex-induced vibrations,'' \emph{Journal of Fluid Mechanics}, vol. 861,
  pp. 119--137, 2019.

\bibitem{jagtap2020adaptive}
A.~D. Jagtap, K.~Kawaguchi, and G.~E. Karniadakis, ``Adaptive activation
  functions accelerate convergence in deep and physics-informed neural
  networks,'' \emph{Journal of Computational Physics}, vol. 404, p. 109136,
  2020.

\bibitem{hendrycks2016gaussian}
D.~Hendrycks and K.~Gimpel, ``{Gaussian error linear units (GELUs)},''
  \emph{arXiv preprint arXiv:1606.08415}, 2016.

\bibitem{deepreach}
S.~Bansal and C.~J. Tomlin, ``{DeepReach}: A deep learning approach to
  high-dimensional reachability,'' in \emph{2021 IEEE International Conference
  on Robotics and Automation (ICRA)}.\hskip 1em plus 0.5em minus 0.4em\relax
  IEEE, 2021, pp. 1817--1824.

\bibitem{zhang2023approximating}
L.~Zhang, M.~Ghimire, W.~Zhang, Z.~Xu, and Y.~Ren, ``{Approximating
  discontinuous Nash equilibrial values of two-player general-sum differential
  games},'' in \emph{2023 IEEE International Conference on Robotics and
  Automation (ICRA)}.\hskip 1em plus 0.5em minus 0.4em\relax IEEE, 2023, pp.
  3022--3028.

\bibitem{bui2022optimizeddp}
M.~Bui, G.~Giovanis, M.~Chen, and A.~Shriraman, ``Optimizeddp: An efficient,
  user-friendly library for optimal control and dynamic programming,''
  \emph{arXiv preprint arXiv:2204.05520}, 2022.

\bibitem{crandall1983viscosity}
M.~G. Crandall and P.-L. Lions, ``{Viscosity solutions of Hamilton-Jacobi
  equations},'' \emph{Transactions of the American mathematical society}, vol.
  277, no.~1, pp. 1--42, 1983.

\bibitem{osher1991high}
S.~Osher and C.-W. Shu, ``{High-order essentially nonoscillatory schemes for
  Hamilton--Jacobi equations},'' \emph{SIAM Journal on numerical analysis},
  vol.~28, no.~4, pp. 907--922, 1991.

\bibitem{osher2004level}
S.~Osher, R.~Fedkiw, and K.~Piechor, ``Level set methods and dynamic implicit
  surfaces,'' \emph{Appl. Mech. Rev.}, vol.~57, no.~3, pp. B15--B15, 2004.

\bibitem{mitchell2005toolbox}
I.~M. Mitchell and J.~A. Templeton, ``{A toolbox of Hamilton-Jacobi solvers for
  analysis of nondeterministic continuous and hybrid systems},'' in
  \emph{Hybrid Systems: Computation and Control: 8th International Workshop,
  HSCC 2005, Zurich, Switzerland, March 9-11, 2005. Proceedings 8}.\hskip 1em
  plus 0.5em minus 0.4em\relax Springer, 2005, pp. 480--494.

\bibitem{mitchell2003}
I.~M. Mitchell and C.~J. Tomlin, ``{Overapproximating reachable sets by
  Hamilton-Jacobi projections},'' \emph{journal of Scientific Computing},
  vol.~19, no.~1, pp. 323--346, 2003.

\bibitem{han2020convergence}
J.~Han and J.~Long, ``{Convergence of the deep BSDE method for coupled
  FBSDEs},'' \emph{Probability, Uncertainty and Quantitative Risk}, vol.~5,
  no.~1, pp. 1--33, 2020.

\bibitem{han2018solving}
J.~Han, A.~Jentzen, and W.~E, ``Solving high-dimensional partial differential
  equations using deep learning,'' \emph{Proceedings of the National Academy of
  Sciences}, vol. 115, no.~34, pp. 8505--8510, 2018.

\bibitem{nakamura2021adaptive}
T.~Nakamura-Zimmerer, Q.~Gong, and W.~Kang, ``{Adaptive deep learning for
  high-dimensional Hamilton--Jacobi--Bellman equations},'' \emph{SIAM Journal
  on Scientific Computing}, vol.~43, no.~2, pp. A1221--A1247, 2021.

\bibitem{fridovich2020efficient}
D.~Fridovich-Keil, E.~Ratner, L.~Peters, A.~D. Dragan, and C.~J. Tomlin,
  ``Efficient iterative linear-quadratic approximations for nonlinear
  multi-player general-sum differential games,'' in \emph{2020 IEEE
  international conference on robotics and automation (ICRA)}.\hskip 1em plus
  0.5em minus 0.4em\relax IEEE, 2020, pp. 1475--1481.

\bibitem{foerster_learning_2017}
J.~N. Foerster, R.~Y. Chen, M.~Al-Shedivat, S.~Whiteson, P.~Abbeel, and
  I.~Mordatch, ``Learning with opponent-learning awareness,''
  \emph{arXiv:1709.04326 [cs]}, Sep. 2017, arXiv: 1709.04326.

\bibitem{sadigh_planning_2018}
D.~Sadigh, N.~Landolfi, S.~S. Sastry, S.~A. Seshia, and A.~D. Dragan,
  ``\BIBforeignlanguage{en}{Planning for cars that coordinate with people:
  leveraging effects on human actions for planning and active information
  gathering over human internal state},''
  \emph{\BIBforeignlanguage{en}{Autonomous Robots}}, vol.~42, no.~7, pp.
  1405--1426, Oct. 2018.

\bibitem{kwon2020humans}
M.~Kwon, E.~Biyik, A.~Talati, K.~Bhasin, D.~P. Losey, and D.~Sadigh, ``When
  humans aren't optimal: Robots that collaborate with risk-aware humans,'' in
  \emph{Proceedings of the 2020 ACM/IEEE International Conference on
  Human-Robot Interaction}, 2020, pp. 43--52.

\bibitem{schwarting2019social}
W.~Schwarting, A.~Pierson, J.~Alonso-Mora, S.~Karaman, and D.~Rus, ``Social
  behavior for autonomous vehicles,'' \emph{Proceedings of the National Academy
  of Sciences}, vol. 116, no.~50, pp. 24\,972--24\,978, 2019.

\bibitem{zahedi2022seeking}
Z.~Zahedi, A.~Khayatian, M.~M. Arefi, and S.~Yin, ``{Seeking Nash equilibrium
  in non-cooperative differential games},'' \emph{Journal of Vibration and
  Control}, p. 10775463221122120, 2022.

\bibitem{li2022off}
J.~Li, Z.~Xiao, J.~Fan, T.~Chai, and F.~L. Lewis, ``{Off-policy Q-learning:
  Solving Nash equilibrium of multi-player games with network-induced delay and
  unmeasured state},'' \emph{Automatica}, vol. 136, p. 110076, 2022.

\bibitem{nikolaidis2017human}
S.~Nikolaidis, D.~Hsu, and S.~Srinivasa, ``Human-robot mutual adaptation in
  collaborative tasks: Models and experiments,'' \emph{The International
  Journal of Robotics Research}, vol.~36, no. 5-7, pp. 618--634, 2017.

\bibitem{sun2018probabilistic}
L.~Sun, W.~Zhan, and M.~Tomizuka, ``Probabilistic prediction of interactive
  driving behavior via hierarchical inverse reinforcement learning,'' in
  \emph{2018 21st International Conference on Intelligent Transportation
  Systems (ITSC)}.\hskip 1em plus 0.5em minus 0.4em\relax IEEE, 2018, pp.
  2111--2117.

\bibitem{peng2019bayesian}
C.~Peng and M.~Tomizuka, ``Bayesian persuasive driving,'' in \emph{2019
  American Control Conference (ACC)}.\hskip 1em plus 0.5em minus 0.4em\relax
  IEEE, 2019, pp. 723--729.

\bibitem{wang2019enabling}
Y.~Wang, Y.~Ren, S.~Elliott, and W.~Zhang, ``Enabling courteous vehicle
  interactions through game-based and dynamics-aware intent inference,''
  \emph{IEEE Transactions on Intelligent Vehicles}, vol.~5, no.~2, pp.
  217--228, 2019.

\bibitem{fridovich2020confidence}
D.~Fridovich-Keil, A.~Bajcsy, J.~F. Fisac, S.~L. Herbert, S.~Wang, A.~D.
  Dragan, and C.~J. Tomlin, ``Confidence-aware motion prediction for real-time
  collision avoidance,'' \emph{The International Journal of Robotics Research},
  vol.~39, no. 2-3, pp. 250--265, 2020.

\bibitem{hu2023learning}
H.~Hu, Z.~Zhang, K.~Nakamura, A.~Bajcsy, and J.~F. Fisac, ``Learning-aware
  safety for interactive autonomy,'' \emph{arXiv preprint arXiv:2309.01267},
  2023.

\bibitem{aumann}
R.~J. Aumann, M.~Maschler, and R.~E. Stearns, \emph{Repeated games with
  incomplete information}.\hskip 1em plus 0.5em minus 0.4em\relax MIT press,
  1995.

\bibitem{cardaliaguet2007differential}
P.~Cardaliaguet, ``Differential games with asymmetric information,'' \emph{SIAM
  journal on Control and Optimization}, vol.~46, no.~3, pp. 816--838, 2007.

\bibitem{cardaliaguet2009numerical}
{Cardaliaguet, Pierre}, ``Numerical approximation and optimal strategies for
  differential games with lack of information on one side,'' \emph{Advances in
  Dynamic Games and Their Applications: Analytical and Numerical Developments},
  pp. 1--18, 2009.

\bibitem{cardaliaguet2012games}
P.~Cardaliaguet and C.~Rainer, ``Games with incomplete information in
  continuous time and for continuous types,'' \emph{Dynamic Games and
  Applications}, vol.~2, no.~2, pp. 206--227, 2012.

\bibitem{starr1969nonzero}
A.~W. Starr and Y.-C. Ho, ``Nonzero-sum differential games,'' \emph{Journal of
  optimization theory and applications}, vol.~3, no.~3, pp. 184--206, 1969.

\bibitem{singarcs}
E.~Cristiani and P.~Martinon, ``{Initialization of the shooting method via the
  Hamilton-Jacobi-Bellman approach},'' \emph{Journal of Optimization Theory and
  Applications}, vol. 146, no.~2, pp. 321--346, 2010.

\bibitem{kumar2020conservative}
A.~Kumar, A.~Zhou, G.~Tucker, and S.~Levine, ``{Conservative Q-learning for
  offline Reinforcement Learning},'' \emph{Advances in Neural Information
  Processing Systems}, vol.~33, pp. 1179--1191, 2020.

\bibitem{petersen2018optimal}
P.~Petersen and F.~Voigtlaender, ``{Optimal approximation of piecewise smooth
  functions using deep ReLU neural networks},'' \emph{Neural Networks}, vol.
  108, pp. 296--330, 2018.

\bibitem{siren}
V.~Sitzmann, J.~Martel, A.~Bergman, D.~Lindell, and G.~Wetzstein, ``Implicit
  neural representations with periodic activation functions,'' \emph{Advances
  in Neural Information Processing Systems}, vol.~33, pp. 7462--7473, 2020.

\bibitem{harsanyi1967games}
J.~C. Harsanyi, ``{Games with incomplete information played by “Bayesian”
  players, I--III Part I. The basic model},'' \emph{Management science},
  vol.~14, no.~3, pp. 159--182, 1967.

\bibitem{chen2021shall}
Y.~Chen, L.~Zhang, T.~Merry, S.~Amatya, W.~Zhang, and Y.~Ren, ``{When shall I
  be empathetic? the utility of empathetic parameter estimation in multi-agent
  interactions},'' in \emph{2021 IEEE International Conference on Robotics and
  Automation (ICRA)}.\hskip 1em plus 0.5em minus 0.4em\relax IEEE, 2021, pp.
  2761--2767.

\bibitem{huang2019reachnn}
C.~Huang, J.~Fan, W.~Li, X.~Chen, and Q.~Zhu, ``Reachnn: Reachability analysis
  of neural-network controlled systems,'' \emph{ACM Transactions on Embedded
  Computing Systems (TECS)}, vol.~18, no.~5s, pp. 1--22, 2019.

\bibitem{manzanas2022reachability}
D.~Manzanas~Lopez, P.~Musau, N.~P. Hamilton, and T.~T. Johnson, ``Reachability
  analysis of a general class of neural ordinary differential equations,'' in
  \emph{International Conference on Formal Modeling and Analysis of Timed
  Systems}.\hskip 1em plus 0.5em minus 0.4em\relax Springer, 2022, pp.
  258--277.

\bibitem{hu2020reach}
H.~Hu, M.~Fazlyab, M.~Morari, and G.~J. Pappas, ``Reach-sdp: Reachability
  analysis of closed-loop systems with neural network controllers via
  semidefinite programming,'' in \emph{2020 59th IEEE conference on decision
  and control (CDC)}.\hskip 1em plus 0.5em minus 0.4em\relax IEEE, 2020, pp.
  5929--5934.

\bibitem{everett2021reachability}
M.~Everett, G.~Habibi, C.~Sun, and J.~P. How, ``Reachability analysis of neural
  feedback loops,'' \emph{IEEE Access}, vol.~9, pp. 163\,938--163\,953, 2021.

\bibitem{entesari2023automated}
T.~Entesari and M.~Fazlyab, ``Automated reachability analysis of neural
  network-controlled systems via adaptive polytopes,'' in \emph{Learning for
  Dynamics and Control Conference}.\hskip 1em plus 0.5em minus 0.4em\relax
  PMLR, 2023, pp. 407--419.

\bibitem{lin2023generating}
A.~Lin and S.~Bansal, ``Generating formal safety assurances for
  high-dimensional reachability,'' in \emph{2023 IEEE International Conference
  on Robotics and Automation (ICRA)}.\hskip 1em plus 0.5em minus 0.4em\relax
  IEEE, 2023, pp. 10\,525--10\,531.

\bibitem{hsu2023safety}
K.-C. Hsu, H.~Hu, and J.~F. Fisac, ``The safety filter: A unified view of
  safety-critical control in autonomous systems,'' \emph{Annual Review of
  Control, Robotics, and Autonomous Systems}, vol.~7, 2023.

\bibitem{bressan2010noncooperative}
A.~Bressan, ``Noncooperative differential games. a tutorial,'' \emph{Department
  of Mathematics, Penn State University}, vol.~81, 2010.

\bibitem{evans2022partial}
L.~C. Evans, \emph{Partial differential equations}.\hskip 1em plus 0.5em minus
  0.4em\relax American Mathematical Society, 2022, vol.~19.

\end{thebibliography}
\bibliographystyle{IEEEtran}

\newpage
\section{Biography Section}
\vspace{-0.3in}
\begin{IEEEbiography}[{\includegraphics[width=1in,height=1.25in,clip,keepaspectratio]{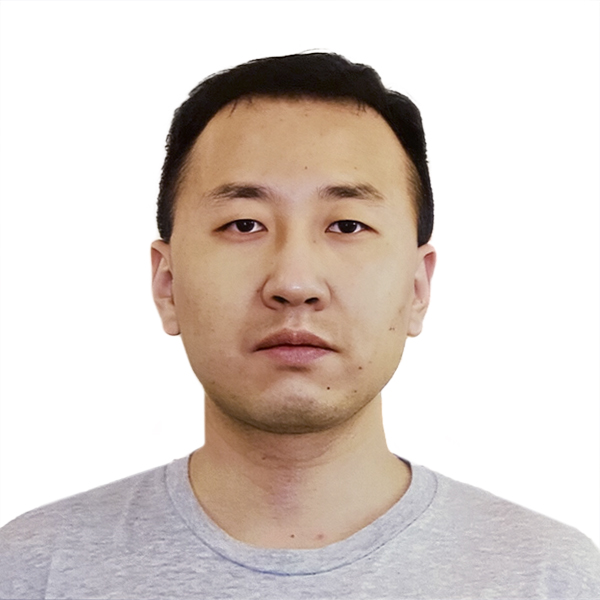}}]{Lei Zhang} received his B.Eng. degree in process equipment and control engineering and M.Sci. degree in material engineering from Xi'an Jiaotong University, China, in 2010 and 2012, respectively. He is a Ph.D. candidate and research assistant in Design Informatics Lab, Arizona State University. His research interests include Machine Learning, Optimization, and Game Theory which are applied to human-robot interactions.
\end{IEEEbiography}

\vspace{-0.3in}
\begin{IEEEbiography}[{\includegraphics[width=1in,height=1.25in,clip,keepaspectratio]{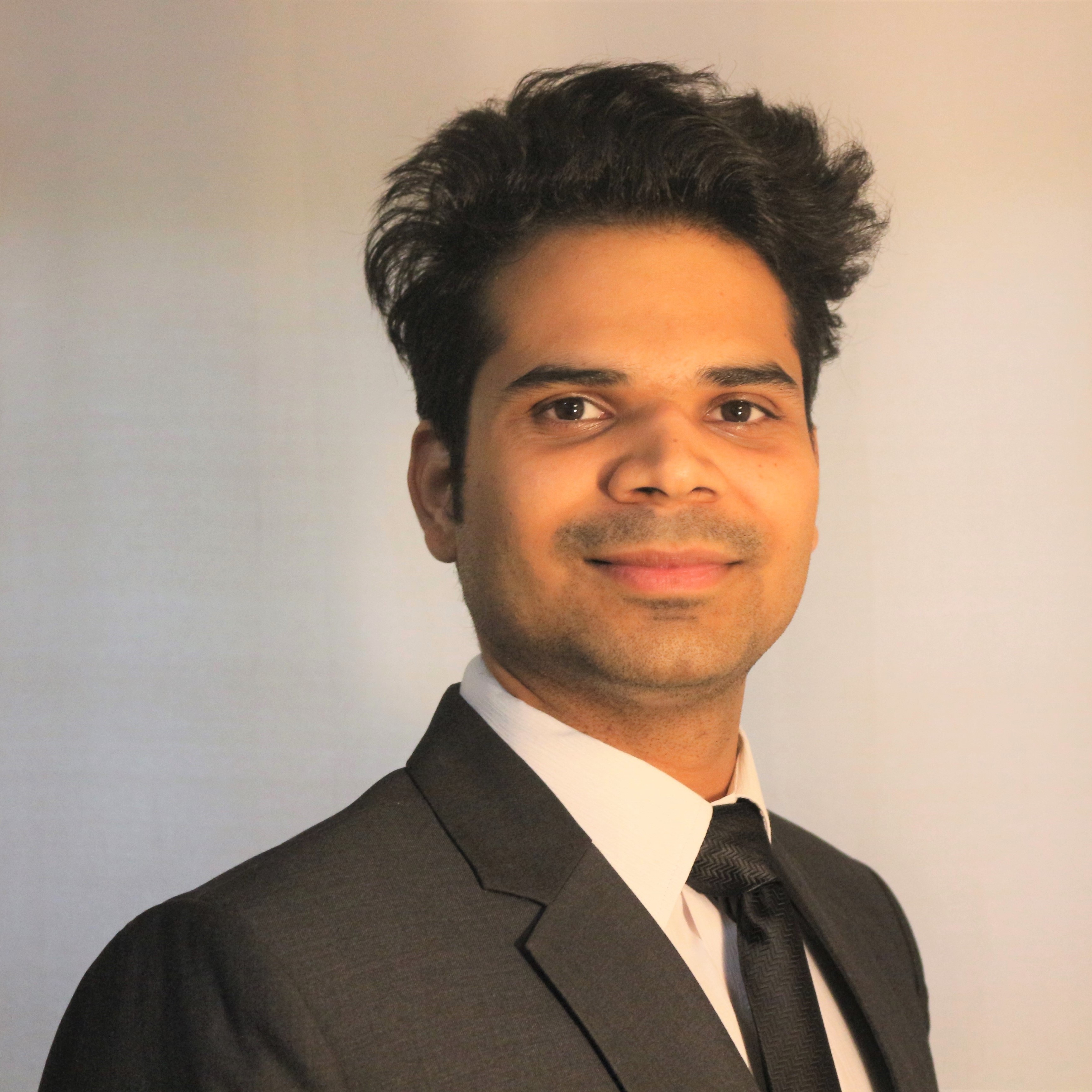}}]{Mukesh Ghimire} received his B.S. in Mechanical Engineering with minors in Computer Science and Mathematics from the University of Mississippi in 2021. He is a Ph.D. student and research assistant in the Design Informatics Lab, Arizona State University. His research interests include Game Theory, Artificial Intelligence and Reinforcement Learning. \end{IEEEbiography}

\vspace{-0.3in}
\begin{IEEEbiography}[{\includegraphics[width=1in,height=1.25in,clip,keepaspectratio]{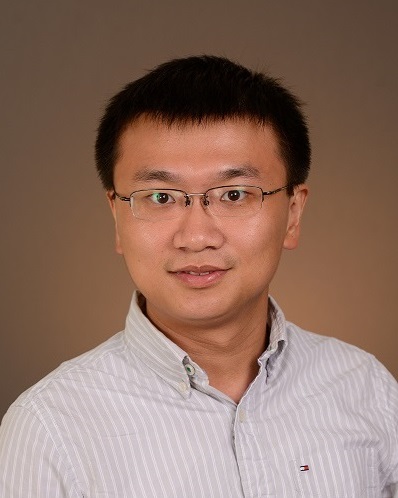}}]{Wenlong Zhang} received the B.Eng. degree (Hons.) in control science and engineering from Harbin Institute of Technology, Harbin, China, in 2010, and the M.S. degree in mechanical engineering in 2012, the M.A. degree in statistics in 2013, and Ph.D. degree in mechanical engineering in 2015, all from the University of California, Berkeley, CA, USA. He is currently an Associate Professor in the School of Manufacturing Systems and Networks at ASU, where he directs the ASU Robotics and Intelligent Systems (RISE) Lab.
His research interests include dynamic systems and control, interactive robotics, and human-machine collaboration. 
\end{IEEEbiography}

\vspace{-0.3in}
\begin{IEEEbiography}[{\includegraphics[width=1in,height=1.25in,clip,keepaspectratio]{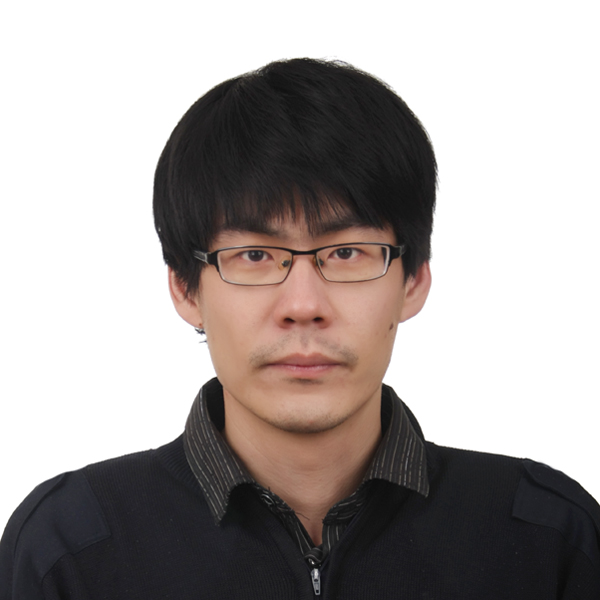}}]{Zhe Xu} received the B.S. and M.S. degrees in Electrical Engineering from Tianjin University, Tianjin, China, in 2011 and 2014, respectively. He received the Ph.D. degree in Electrical Engineering at Rensselaer Polytechnic Institute, Troy, NY, in 2018. He is currently an assistant professor in the School for Engineering of Matter, Transport, and Energy at Arizona State University. Before joining ASU, he was a postdoctoral researcher at the Oden Institute for Computational Engineering and Sciences at the University of Texas at Austin, Austin, TX. His research interests include formal methods, autonomous systems, control systems, and reinforcement learning.
\end{IEEEbiography}

\vspace{-0.3in}
\begin{IEEEbiography}[{\includegraphics[width=1in,height=1.25in,clip,keepaspectratio]{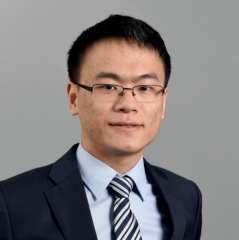}}]{Yi Ren} is an Associate Professor in Mechanical Engineering at Arizona State University. He received his Ph.D. in Mechanical Engineering from the University of Michigan Ann Arbor in 2012 and his B.Eng. degree in Automotive Engineering from Tsinghua University in 2007. From 2012 to 2014 he was a post-doctoral researcher at the University of Michigan. Dr. Ren’s current research focuses on safety in AI-enabled engineering systems. He won the Best Paper Award at the 2015 ASME International Design and Engineering Technical Conferences (IDETC).
\end{IEEEbiography}

\vfill

\end{document}